\documentclass[11pt]{article}

\usepackage[final]{acl}
\usepackage{times}
\usepackage{latexsym}
\usepackage[T1]{fontenc}
\usepackage[utf8]{inputenc}
\usepackage{microtype}
\usepackage{inconsolata}
\usepackage{graphicx}
\usepackage{amsmath}
\usepackage{longtable}
\usepackage{placeins}
\usepackage{float}
\usepackage{comment}

\title{Where meaning lives: Layer-wise accessibility of psycholinguistic features in encoder and decoder language models}

\author{
  \textbf{Taisiia Tikhomirova\textsuperscript{1,2}},
  \textbf{Dirk U. Wulff\textsuperscript{1,3}}
\\
\\
  \textsuperscript{1}Max Planck Institute for Human Development, Berlin, Germany \\
  \textsuperscript{2}Technische Universität Berlin, Berlin, Germany \\
  \textsuperscript{3}Department of Psychology, University of Basel, Basel, Switzerland
\\
\\
  \small{
    \textbf{Correspondence:} \href{mailto:tikhomirova@mpib-berlin.mpg.de}{tikhomirova@mpib-berlin.mpg.de}
  }
}

\begin{document}
\maketitle

\begin{abstract}
Understanding where transformer language models encode psychologically meaningful aspects of meaning is essential for both theory and practice. We conduct a systematic layer-wise probing study of 58 psycholinguistic features across 10 transformer models, spanning encoder-only and decoder-only architectures, and compare three embedding extraction methods. We find that apparent localization of meaning is strongly method-dependent: contextualized embeddings yield higher feature-specific selectivity and different layer-wise profiles than isolated embeddings. Across models and methods, final-layer representations are rarely optimal for recovering psycholinguistic information with linear probes. Despite these differences, models exhibit a shared depth ordering of meaning dimensions, with lexical properties peaking earlier and experiential and affective dimensions peaking later. Together, these results show that where meaning “lives” in transformer models reflects an interaction between methodological choices and architectural constraints.

\end{abstract}

\section{Introduction}
How do language models represent meaningful dimensions of language, such as emotion, concreteness, or sensory experience? As transformer-based models become increasingly integrated into real-world systems, understanding the internal dimensions they encode is essential. This matters both theoretically, for assessing whether language models reflect human semantic representations, and for practice, where interpretability and transparency are critical for safe deployment in meaning-sensitive domains such as education, mental health, and human–AI interaction.

\begin{table*}[t!]
  \centering
  \begin{tabular}{lccc}
    \hline
    \textbf{Encoder Model} & \textbf{Params} & \textbf{$L$} & \textbf{$d_{\text{model}}$} \\
    \hline
    BERT Large       & 336M & 24 & 1024 \\
    RoBERTa Large     & 355M & 24 & 1024 \\
    DeBERTa-v3 Large & 304M & 24 & 1024 \\
    BGE-M3           & 567M & 24 & 1024 \\
    Jina-v3          & 570M & 24 & 1024 \\
    \hline
  \end{tabular}
  \quad
  \begin{tabular}{lccc}
    \hline
    \textbf{Decoder Model} & \textbf{Params} & \textbf{$L$} & \textbf{$d_{\text{model}}$}\\
    \hline
    Mistral-24B      & 24B & 40 & 5120 \\
    Phi-4           & 14B & 40 & 5120 \\
    GPT-OSS-20B     & 20B & 24 & 2880 \\
    Gemma-3-27B     & 27B & 62 & 5376 \\
    Qwen3-32B       & 32B & 64 & 5120 \\
    \hline
  \end{tabular}
  \caption{\textbf{Model Specifications.} Overview of the ten transformer models evaluated, separated by architectural class. $L$ denotes the number of layers, and $d$ denotes the hidden dimension size.}
  \label{tab:model_specs}
\end{table*}

Transformer-based models produce contextual token embeddings at each layer, which have been widely used to study the distribution of linguistic information within the models. Prior probing work has reported systematic differences across model layers, suggesting that surface, syntactic, and semantic features may be preferentially represented at different depths \citep{Jawahar_2019, Tenney_2019, Hewitt_2019, lin-etal-2019-open}. These findings have influenced both interpretive claims about transformer representations and common modeling practices, including the frequent use of final-layer embeddings for semantic tasks.

Recent work has begun to challenge the default reliance on final-layer representations by showing that internal layers are often more informative for downstream tasks and that certain meaning dimensions—most notably emotion—are most decodable in middle layers \citep{skean2025layerlayeruncoveringhidden, zhang2025decodingemotiondeepsystematic}. At the same time, layer-wise localization of such features appears to depend strongly on model architecture \citep{Liu_2024}. Because existing evidence largely comes from single models or narrow sets of dimensions, it remains unclear whether reported patterns reflect general principles or model- and dimension-specific effects. This uncertainty is particularly consequential given growing interest in whether language models encode psychologically meaningful dimensions of language \citep{Waldis_2024, Zhu_2024, Xu_2025}. Meaning is inherently multi-faceted, encompassing not only affective properties but also perceptual, cognitive, and social aspects. A systematic, cross-architectural investigation spanning a broader range of meaning dimensions is therefore needed to determine how meaning is distributed and transformed across layers.

Interpreting hidden representations also raises unresolved methodological challenges. Because language models encode word meaning in context, the choice of context used for embedding extraction is itself a critical design decision. Prior studies vary widely: some embed target words in fixed templates (e.g., “What is the meaning of WORD?”) \citep{liétard2021languagemodelsknowway, Petroni2019LanguageMA}, while others rely on naturally occurring sentences, from single instances \citep{Chang2019WhatDT} or averaged across contexts \citep{Bommasani2020InterpretingPC}. Most work adopts only one strategy. If apparent layer-wise localization depends on extraction method, conclusions about how meaning is distributed in language models may be method-dependent rather than intrinsic to the models.

These considerations motivate a systematic analysis that jointly considers a broad range of meaning dimensions, model architectures, and embedding extraction methods. Our study analyzes how meaning dimensions in the form of 58 psycholinguistic features \citep{Wulff_2024} are encoded across layers in 10 transformer models spanning two architectural classes. To assess methodological stability, we apply three embedding extraction methods and use linear probes to measure the amount of feature-selective information recoverable at each layer.

We make three contributions to the study of meaning representation in language models. First, we show that embedding extraction methods and model architecture affect both recoverable information and layer-wise localization, indicating that conclusions based on a single method may be unstable. Second, we show that final-layer embeddings are rarely optimal for recovering psycholinguistic meaning with linear probes across architectures and extraction methods, implying that defaulting to final-layer representations can miss information that is most accessible in intermediate layers. Finally, we show that models exhibit a depth ordering of meaning dimensions that is shared within and across model architectures, with depth rising from lexical to semantic features.

\begin{table*}[t!]
  \centering
  \begin{tabular}{p{0.17\textwidth} p{0.69\textwidth} r}
    \hline
    \textbf{Category} & \textbf{Description} & \textbf{N} \\
    \hline
    Frequency &
    How often a word occurs in language, based on log-transformed frequency estimates from diverse spoken and written corpora. & 10 \\

    Motor &
    Degree to which a word is associated with bodily actions or motor experiences. & 7 \\

    Sensory &
    Strength of perceptual experience associated with a word across sensory modalities, including visual, auditory, tactile, olfactory, and gustatory experience. & 6 \\

    Semantic Diversity &
    Variability of contexts in which a word appears, reflecting how semantically diverse or context-specific its usage is. & 6 \\

    Visual Lexical Decision &
    Accuracy and response speed in tasks where participants judge whether a visually presented letter string is a real word. & 6 \\

    Familiarity &
    How well a word is known to speakers, including when it is learned, how many individuals recognize or understand it, and how frequently it is encountered in language use. & 4 \\

    Auditory Lexical Decision &
    Accuracy and response speed in tasks where participants judge whether a spoken stimulus corresponds to a real word. & 4 \\

    Valence &
    Emotional polarity of a word, ranging from negative to positive affective meaning. & 2 \\

    Arousal &
    The degree of emotional intensity or activation elicited by a word. & 2 \\

    Dominance &
    Extent to which a word evokes feelings of control, power, or influence versus submission or passivity. & 2 \\

    Naming &
    Speed and accuracy with which speakers produce a word’s pronunciation when presented with its written or visual form. & 2 \\

    Semantic Decision &
    Accuracy and response latency in tasks where participants judge whether a word refers to something concrete or abstract, based on semantic decision data from the Calgary database. & 2 \\

    Age of Acquisition &
    Estimated age (in years) at which speakers report having learned a word, reflecting the timing of lexical acquisition. & 1 \\

    Concreteness &
    Extent to which a word refers to tangible, perceptible entities as opposed to abstract concepts. & 1 \\

    Semantic Neighborhood &
    Density or similarity of meanings surrounding a word in semantic space, indicating how closely related it is to other words. & 1 \\

    Social / Moral &
    Extent to which a word conveys social, interpersonal, or moral meaning relevant to human interaction and norms. & 1 \\

    Iconicity / Transparency &
    Degree to which a word’s form resembles or transparently conveys its meaning. & 1 \\
    \hline
  \end{tabular}
  \caption{\textbf{Feature Categories.} Overview of the feature categories used in the analysis. N - Number of datasets}
  \label{tab:norm_categories}
\end{table*}

\section{Methodology}

\subsection{Word features}

We relied on the psychNorms metabase \cite{Wulff_2024}, a large-scale aggregation of psycholinguistic features derived from dozens of independent behavioral studies. The database includes human-validated word-level features spanning affective (e.g., valence, arousal), semantic (e.g., concreteness, imageability, semantic diversity), developmental (e.g., age of acquisition), sensory–motor, lexical (e.g., frequency), and behavioral performance measures (e.g., accuracy and response times in lexical decision tasks; see Table~\ref{tab:norm_categories}). With the exception of frequency and semantic diversity, which are corpus-derived, these features are based on Likert-style ratings or task behavior, providing a principled, behaviorally grounded target space for probing representations in language models.

Coverage in psychNorms varies substantially across features, ranging from 703 to 79{,}671 words. To ensure comparability across features while controlling computational cost, we constructed a subset of 9,966 words greedily maximizing overlap among the 58 highest-coverage (N>4600) features. This procedure minimizes confounds arising from differences in word sets and sample sizes across features while retaining broad lexical coverage. The word set has a median rank of 9{,}201 (IQR = [3{,}858, 16{,}260]) out of the total length of 57{,}214  of the SUBTLEXUS word frequency dictionary and contains 52\% nouns, 22\% verbs, 19\% adjectives, and 7\% other parts of speech. 

\begin{figure*}[h!]
    \centering
    \includegraphics[width=.95\textwidth]{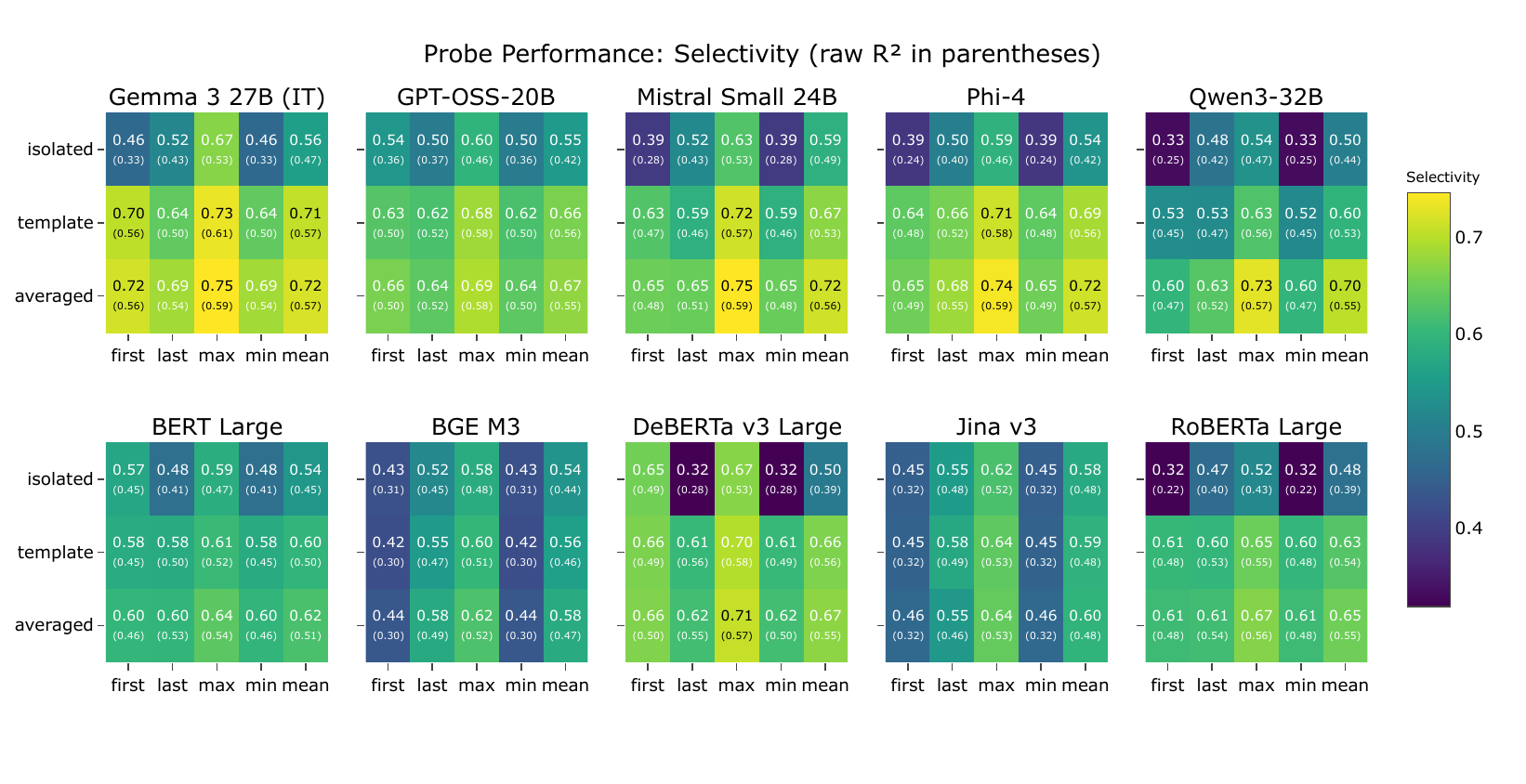}
    \caption{Linear probe performance (selectivity and raw $R^2$ in parenthesis) averaged over features across ten language models. Heatmaps show the predictive power of five embedding extraction methods (Y-axis) summarized over five layer metrics: average over first, last, best and worst performing layers and mean over all layers (X-axis), separated by model architecture (decoders: top; encoders: bottom).}
    \label{fig:extraction}
\end{figure*}

\subsection{Transformer models}

We examined ten openly available transformer models accessed via the HuggingFace platform, spanning both major architectural paradigms: encoder-only and decoder-only models (see Table~\ref{tab:model_specs}). Encoder-based models included BERT Large \cite{devlin2019bertpretrainingdeepbidirectional}, RoBERTa Large \cite{liu2019robertarobustlyoptimizedbert}, DeBERTa-v3 Large \cite{he2023debertav3improvingdebertausing}, BGE-M3 \cite{chen2024bgem3embeddingmultilingualmultifunctionality}, and Jina-v3 \cite{sturua2024jinaembeddingsv3multilingualembeddingstask}. Decoder-based models included Mistral-24B \cite{mistral2025small3}, Phi-4 \cite{abdin2024phi4technicalreport}, Qwen3-32B \cite{yang2025qwen3technicalreport}, GPT-OSS-20B \cite{openai2025gptoss120bgptoss20bmodel}, and Gemma-3-27B \cite{gemmateam2025gemma3technicalreport}.

This selection spans a wide range of parameter scales, pre- and post-training objectives, and architectural choices. By contrasting encoder and decoder architectures within a unified probing framework, we aim to distinguish architectural regularities in the localization of psycholinguistic features from model-specific idiosyncrasies.

\subsection{Embedding extraction}

A central methodological challenge in representational probing is determining the context in which a word embedding should be extracted. We therefore compare three embedding extraction methods used in the literature \cite{gurnee2024languagemodelsrepresentspace, liétard2021languagemodelsknowway, Bommasani2020InterpretingPC, chronis-erk-2020-bishop}: two contextualized extraction methods, called \emph{template} and \emph{averaged}, that differ in the amount and type of contextual information provided, and a baseline extraction method, called \emph{isolated}, providing no context. For each approach, embeddings were extracted at the input layer and at the output of each transformer block (post-norm). For words consisting of multiple tokens, token-level embeddings were averaged.

Formally, let $h_{\ell}(w, c)$ denote the hidden state of word $w$ in context $c$ at layer $\ell$. We define the three extraction methods as:

\begin{equation}
\begin{aligned}
e_{\ell}^{\mathrm{iso}}(w) &= h_{\ell}(w, \emptyset).\\
e_{\ell}^{\mathrm{temp}}(w) &= h_{\ell}(w, s_{\mathrm{temp}}(w))\\
e_{\ell}^{\mathrm{avg}}(w) &= \frac{1}{50} \sum_{i=1}^{50} h_{\ell}(w, s_{wi}).
\end{aligned}
\end{equation}

The context $(s_{\mathrm{temp}}(w))$ in the calculation of the template embedding $e_{\ell}^{\mathrm{temp}}$ is the sentence ``What is the meaning of the word [word]?'', with [word] being replaced by $w$. The context $s_{wi}$ in the calculation of the aggregate embedding $e_{\ell}^{\mathrm{avg}}$ is one of 50 sentences sampled at random from a representative subset of the C4 corpus \cite{Raffel_2020} containing the word $w$. We use 50 contexts as a compromise between stability and computational cost; pilot analyses showed diminishing returns beyond this point.

\begin{figure*}[t!]
    \centering
  \includegraphics[width=.95\textwidth]{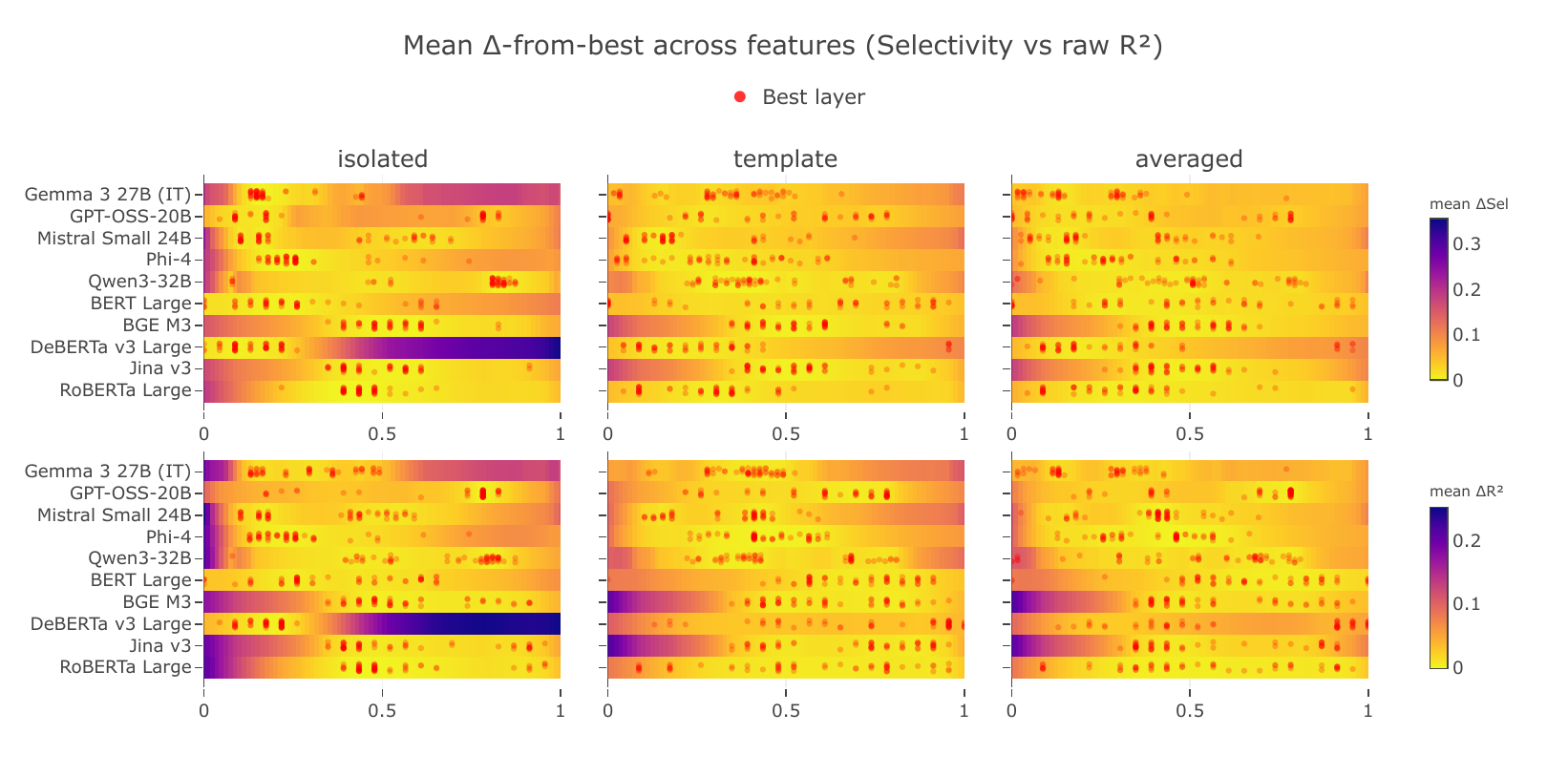}
  \caption{The heatmaps visualize the average performance drop ($\Delta$) relative to the best-performing layer for each model across three embedding extraction methods: isolated, template, and averaged. The top row displays results for selectivity, while the bottom row shows raw $R^2$ scores. Within each panel, decoders are grouped at the top and encoders at the bottom, with the x-axis representing the normalized layer index ($0 = \text{first}$, $1 = \text{last}$). Red dots mark the single best layer (argmax) for each individual model-feature pair.}
  \label{fig:architectures}
\end{figure*}

\subsection{Locating meaning}

To locate psycholinguistic information within transformer models, we applied layer-wise linear probing using ridge regression. We focus on linear probes to assess which information is directly accessible from model representations, rather than to maximize predictive accuracy, as more expressive probes can reconstruct information not explicitly encoded \cite{belinkov-2022-probing}.

Importantly, high decoding performance alone does not guarantee that a representation selectively encodes the target feature. Linear probes may exploit correlations with unrelated lexical or distributional properties, or achieve above-chance performance even under label permutation \cite{ravichander2021probing, hewitt2019designing}.

To control for these confounds, we use selectivity as our primary outcome measure. Selectivity compares decoding performance on the true target to performance on a matched control task with permuted feature labels, isolating information specifically aligned with the feature of interest \cite{hewitt2019designing, belinkov-2022-probing}.

For each combination of psycholinguistic feature, model, layer, and embedding extraction method, we fit a separate ridge regression using embedding vectors as predictors and human feature values as targets. Models were trained using nested 5-fold cross-validation on a random subset of 4{,}000 words, with regularization strength $\alpha \in [1{,}000;10{,}000]$. Performance was evaluated using out-of-sample $R^2_{obs}$. This procedure was repeated ten times, yielding 50 estimates per combination, which were averaged for analysis.

We repeated the same procedure under random permutation of the target feature values to estimate chance-level performance. Permutations were performed separately for each random subset of 4{,}000 words. Performance under permutation ($R^2_{rand}$) ranged between -0.58 and -0.01. Selectivity was then computed as 

\[
R^2_{sel} = R^2_{obs} - R^2_{rand}
\]

Finally, to address localization, we calculate the center of mass as
\[
COM = \frac{\sum^L_{\ell = 1} \lambda(\ell) \Delta R^2_{sel, \ell}}{\sum^L_{\ell = 1} \Delta R^2_{sel, \ell}}
\]

with $\lambda(\ell) = \ell/L$ being the relative layer and $\Delta R^2_{sel, \ell}$ the selectivity relative to the lowest-selectivity layer. Additionally, we report the layers with maximum selectivity.

\begin{figure*}[h!]
    \centering
  \includegraphics[width=.95\textwidth]{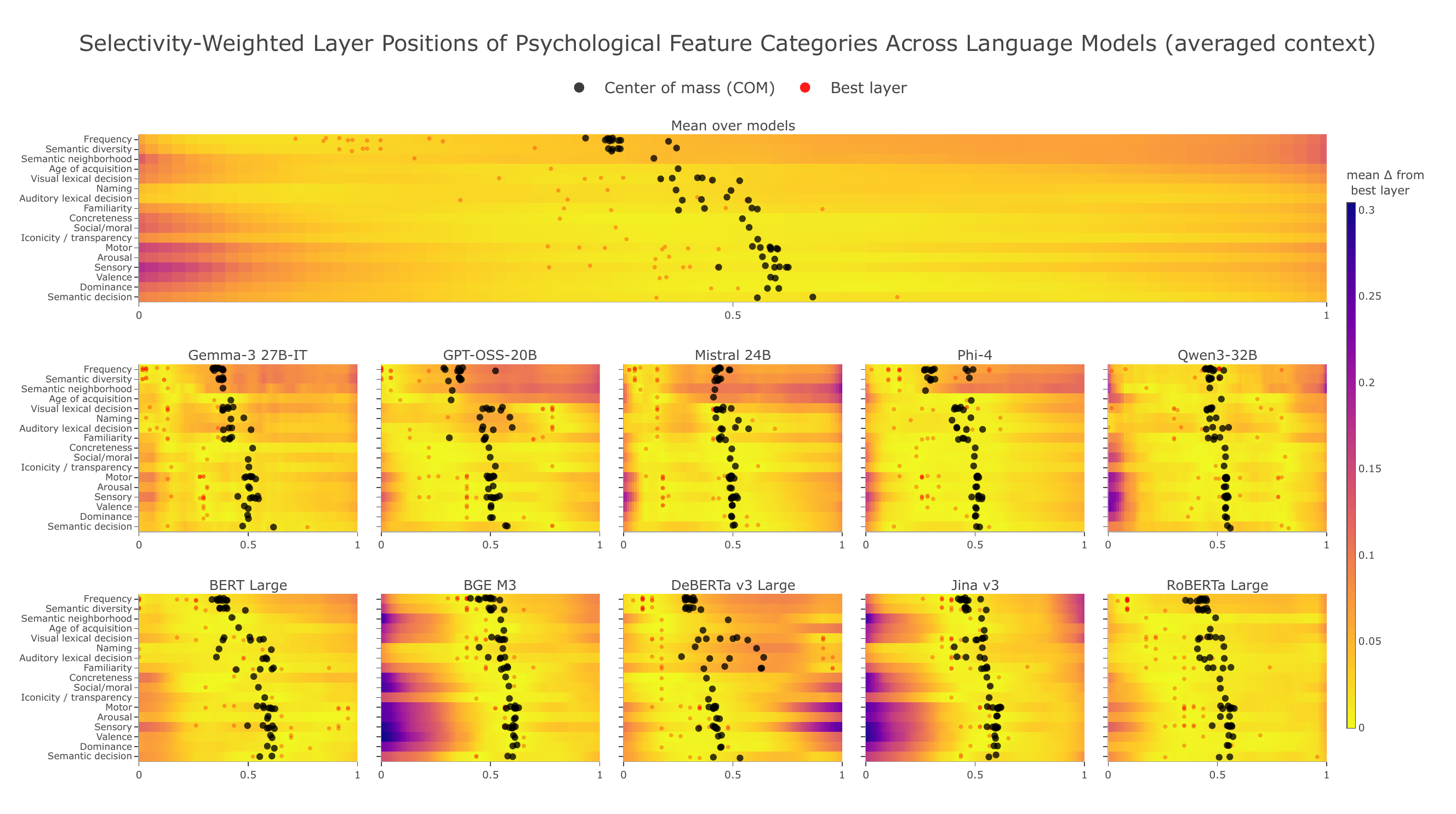}
  \caption{Selectivity-weighted layer positions of psycholinguistic feature categories for the averaged embedding extraction method. The heatmap depicts the mean $\Delta$-from-best-layer across features comprising each category (X-axis: normalized layer index from first to last; Y-axis: psycholinguistic feature categories) based on selectivity score. Black dots indicate the selectivity-weighted center of mass (COM) of each feature’s layer profile, while red dots mark the single best-performing layer (argmax) for each feature. The top panel corresponds to the mean over all models, and the two bottom panels correspond to language models (decoders, top row, encoders, bottom row).}
  \label{fig:features}
\end{figure*}

\section{Results}

We report results from layer-wise linear probing of 58 psycholinguistic features across ten transformer models, focusing on how embedding extraction method affects decodability and localization, architectural differences between encoder and decoder models, and the relative depth at which different psycholinguistic features are most accessible. Unless otherwise stated, results are based on selectivity.  

\subsection{Localization of psycholinguistic features is highly dependent on embedding extraction}

Figure~\ref{fig:extraction} shows that contextualized embeddings, whether template-based or context-averaged, consistently yield higher linear decodability than isolated embeddings. Across all models and features, moving to contextualized extraction increases median selectivity by 0.112 (0.106 raw $R^{2}$), with improvements observed for 100\% of features. Among contextualized methods, averaged yielded consistently higher selectivity (and raw $R^{2}$) than template, suggesting a benefit for using richer and more diverse contexts in the extraction of psycholinguistic information.   

The extraction method also influences the inferred layer-wise profiles: while isolated embeddings exhibit more pronounced variance and later peaks, contextualized embeddings produce flatter profiles, maintaining 80–90\% of peak selectivity throughout large portions of the network.

Critically, final-layer representations are rarely optimal for recovering psycholinguistic information. Across all feature-model combinations, maximal selectivity is never achieved in the final layer; instead, optimal layers are distributed throughout the network depth. Together, these findings indicate that conclusions regarding where meaning is represented are contingent on the extraction strategy. Differences induced by extraction method are comparable in magnitude to differences between models or architectures, indicating that layer-wise analyses that fix a single extraction strategy risk drawing unstable or misleading conclusions.

\begin{figure*}[h!]
    \centering
  \includegraphics[width=.95\textwidth]{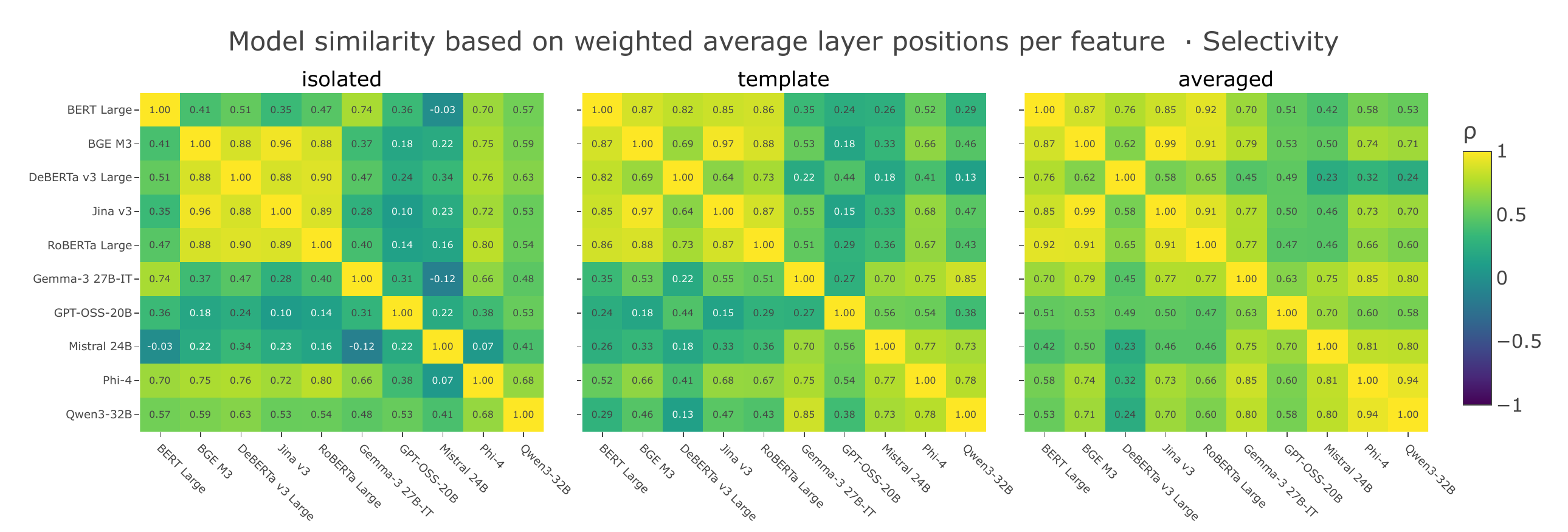}
  \caption{Each panel shows the pairwise Spearman correlation (p) between models, computed from vectors of feature-specific center-of-mass (COM) layer positions within a given embedding extraction method (isolated, template, averaged). For each model and feature, the COM was calculated (using selectivity scores), summarizing where in the network a feature is most strongly represented. Correlations are computed across feature orders, yielding a similarity matrix that reflects how similarly different models localize psycholinguistic information across layers.}
  \label{fig:correlations}
\end{figure*}

\subsection{Models differ in how psycholinguistic features are distributed}

Conditioning on extraction method, models vary in how meaning-related information is concentrated around optimal layers (Figure~\ref{fig:architectures}). Across both encoder and decoder families, models vary substantially in how tightly meaning-related information is concentrated around their best-performing layer. While RoBERTa-Large exhibits broad localization, other models like Qwen3-32B show sharp mid-layer peaks with steep performance drops toward both input and output. These differences are only partially aligned with architectural class: under contextualized extraction, several decoders match encoders in profile breadth, whereas some encoders exhibit pronounced mid-layer concentration. Thus, the degree to which psycholinguistic information is distributed across layers is more model- rather than architecture-dependent.

Figure~\ref{fig:architectures} further illustrates the low selectivity of final-layer and even later-layer representations. This degradation is most pronounced in decoders: the last 20\% of layers comprise only 0.86\% of best-selectivity layers (0.52\% raw $R^2$), with mean selectivity and raw performance drops of 0.07 and 0.051, respectively. In encoders, these layers account for 7.41\% of best-selectivity layers (29.66\% raw $R^2$), showing drops of 0.054 and 0.0267. These results underline that output-layer embeddings substantially underrepresent psycholinguistic information accessible in earlier layers.

\subsection{Psycholinguistic features exhibit a stable depth ordering}

We now shift focus to psycholinguistic features. Showing the results for averaged embeddings, Figure~\ref{fig:features} reveals a consistent selectivity-weighted center of mass (COM) ordering across layers. See Figures 5-15 for the results of other extraction methods and raw $R^2$ for individual features and categories. Lexical and usage properties (e.g., frequency) peak early, while experiential, affective, and social dimensions (e.g., valence, concreteness) peak later, establishing a robust lexico-semantic access ordering.

This ordering is robust across models and architectures. For averaged embeddings, pairwise Spearman correlations between models’ feature-wise COM vectors strictly exceed $\rho .30$ in all comparisons and reach values above $.70$ within architectural classes (see Figure~\ref{fig:correlations}). Importantly, correlations are robust to excluding frequency and semantic diversity features, indicating that these correlations are not driven by objectively determined features (see Figures 17 and 18 in the Appendix). 

Although individual features vary in absolute localization, violations of the overall ordering are rare and unsystematic. No category consistently peaks earlier than all lexical measures or later than all semantic measures, indicating that these results reflect a shared representational ordering rather than model-specific quirks.

\subsection{Shared ordering, distinct realizations across architectures}

Finally, we compare how this shared depth ordering is realized across architectures. Figure 4 shows similarity matrices based on feature-specific COM vectors under different extraction methods. For contextualized embeddings, encoder models form a tight cluster, as do decoder models, yielding a clear block structure. This indicates that while encoders and decoders largely agree on which dimensions emerge earlier versus later, they differ systematically in how this progression is distributed across layers.

This architectural separation is substantially weaker for isolated embeddings, where correlations are noisier and clustering is less pronounced, reinforcing the conclusion that contextualized extraction is necessary to recover stable representational structure. Together, these findings suggest that despite a common ordering, architectural design still plays an important role in how psycholinguistic features are represented. 

\section{Discussion}

This study provides the most comprehensive investigation to date of how psychologically meaningful linguistic dimensions are represented across layers of transformer models. By systematically crossing 58 human-derived semantic features, 10 models spanning both major architectural families, three embedding extraction methods, and full layer-wise probing, we go beyond prior work, which is typically limited to fewer models, narrow feature sets, or fixed extraction strategies. Our findings reveal that conclusions about where psycholinguistic information is represented in language models depend jointly on methodological choices and architectural constraints, and uncover a lexico-semantic depth ordering that generalizes across contemporary transformer models.

\subsection{Embedding extraction is a first-order methodological choice}

Contextualized embeddings (template or averaged) yield substantially higher linear selectivity than isolated word embeddings across all models and feature categories. Moreover, isolated embeddings exhibit sharper peaks and stronger apparent localization; contextualized extraction reveals broader accessibility profiles. These findings indicate that psycholinguistic dimensions rely on contextual processing to become linearly accessible to probes. Such results align with evidence that contextualized representations cannot be reduced to static embeddings without semantic loss \cite{ethayarajh2019contextual, Bommasani2020InterpretingPC} and with distributed accounts of semantic processing that emphasize context-dependent activation \cite{elman2004alternative, wulff2019new}.

Crucially, our findings extend these observations beyond a narrow set of semantic properties to a broad range of psychologically grounded features, including affective, sensory, motor, and social dimensions. At the same time, we emphasize that lower decodability from isolated embeddings does not imply the absence of such information, but rather reduced linear accessibility.

Notably, context-averaged embeddings yield systematically higher selectivity than template-based embeddings. However, the differences may not justify the increased computational cost. Averaging across 50 naturally occurring contexts requires extensive corpus retrieval and repeated inference, whereas a single, minimal template (“What is the meaning of the word [X]?”) suffices to elicit similar levels of feature-specific information. For large-scale probing studies, template-based extraction may therefore offer a computationally efficient alternative to context averaging with minimal loss in linear recoverability.

\subsection{Architecture shapes representational organization}

Encoder and decoder models exhibit systematically different layer-wise accessibility profiles, though the distinction is often subtle. Encoder models tend toward slightly broader distributions , while decoders more frequently concentrate selectivity in intermediate layers with steeper declines toward the input and output.

This architectural contrast is most evident under contextualized extraction and is, however, secondary to model-specific variation. Since several decoders exhibit profiles comparable in breadth to encoders, and some encoders show pronounced mid-layer concentration, architecture constrains but does not uniquely determine representational spread.

These results may help reconcile apparently conflicting findings in the literature. Studies reporting graded, pipeline-like progressions from surface to semantic features have primarily examined encoder models \cite{Jawahar_2019, Tenney_2019}, whereas more recent work identifying mid-layer semantic peaks has focused on decoder-only architectures \cite{Liu_2024, skean2025layerlayeruncoveringhidden}. Our results suggest that both patterns are valid within their respective architectural contexts, rather than mutually contradictory.

\subsection{Final layers underrepresent psycholinguistic information}

Across all models and embedding extraction methods, final-layer representations are rarely optimal for recovering psycholinguistic features via linear probes. These findings challenge the widespread practice of defaulting to final-layer embeddings for semantic analysis. Final layers are optimized for masked- or next-token prediction and downstream task objectives, and their representations may therefore transform information in ways that reduce linear accessibility. Our results extend prior observations regarding final-layer anisotropy and reduced interpretability \cite{ethayarajh2019contextual, skean2025layerlayeruncoveringhidden} to a broad set of psycholinguistic features and across both encoder and decoder architectures.

Importantly, reduced linear decodability should not be interpreted as evidence that psycholinguistic information is lost or absent in final layers. Rather, it indicates that such information is less directly accessible to simple linear readouts, reinforcing the value of layer-wise analyses when probing model representations.

\subsection{A shared ordering of psycholinguistic accessibility}

Despite differences in how information is distributed across layers, models exhibit a remarkably consistent relative lexico-semantic ordering of psycholinguistic features with respect to layer depth, robust across models, with strong rank correlations of feature-specific center-of-mass vectors within models of the same architecture. This consistency indicates that transformer models broadly agree on which types of psycholinguistic information become more accessible earlier versus later in processing, even though they differ in how this progression is distributed across layers. We emphasize that shared ordering does not imply shared representational geometry: models realize this progression in systematically different ways.

\section{Limitations}

\textbf{Language coverage.} All features and probing experiments are restricted to English. Psycholinguistic features such as valence, concreteness, or social meaning may be realized differently across languages due to cultural, lexical, and morphological variation. The architectural patterns observed here may therefore not generalize to multilingual settings. Extending our approach to multilingual features and representations is an important direction for future work.

\textbf{Linear probing as an access measure.} We rely exclusively on linear probes to assess which psycholinguistic features are directly accessible from model representations. This choice follows established recommendations to avoid overinterpreting expressive probes \cite{hewitt2019designing, belinkov-2022-probing}, but it necessarily limits our conclusions. Some dimensions may be encoded in nonlinear or highly distributed forms that linear probes cannot recover. Differences in decodability should therefore be interpreted as differences in accessibility, not as the presence or absence of information.

\textbf{Representations versus behavior.} Our analysis focuses on internal representations rather than model behavior. Higher decodability of a psycholinguistic feature does not guarantee that a model will reliably use or express that information during generation or downstream tasks. Bridging representational analyses with controlled behavioral interventions remains an open challenge for future work.

\textbf{Correlated feature and construct redundancy.} Many psycholinguistic features are correlated (e.g., frequency, age of acquisition, familiarity, and lexical decision measures), reflecting shared behavioral and distributional factors. While selectivity mitigates generic predictability effects, our analyses do not fully disentangle overlapping constructs. Accordingly, the reported layer-wise patterns should be interpreted as reflecting relative accessibility of correlated meaning dimensions rather than sharply separable psycholinguistic modules, even if the demonstrated selectivity ordering implies meaningful differences in feature categories.  

\textbf{Training regime and instruction tuning.} The models in our study differ not only in architecture but also in post-training objectives. We cannot cleanly separate architectural effects from training influences. Future work systematically varying training objectives will be needed to disentangle these factors.

Together, these limitations delineate the scope of our claims. Our results characterize where psycholinguistic information is linearly accessible within model representations under controlled probing conditions, providing a foundation for future work linking internal organization to multilingual generalization and observable behavior.


\bibliography{custom}

\appendix

\appendix
\section{Appendix}
\label{sec:appendix}

This appendix provides supporting resources and full results that complement the main text. Definitions of all psycholinguistic features and their category assignments are provided in Table 3 (based on the psychNorms metabase \cite{Wulff_2024}). We further report additional visualization sets of layer-localization patterns across all models and embedding extraction contexts that are not shown in the main paper.

Figures 5–10 present feature-wise heatmaps of the $\Delta$-from-best-layer score over normalized layer position for each model and context. Black dots indicate the score-weighted center of mass (COM), and red dots mark the single best-performing layer.

Figures 11–15 show the corresponding category-level heatmaps, in which values are averaged across features within each category. While the main text focuses on a single embedding extraction method evaluated using the selectivity score for clarity, the appendix includes the full set of category-level visualizations across embedding extraction methods and for both selectivity and raw $R^2$ scores.

Figure 16 reports pairwise model similarity as a correlation matrix computed from feature-specific COM layer positions (raw $R^2$-weighted), summarizing the extent to which different models localize psycholinguistic information similarly across layers.

Finally, Figures 17–18 report the same pairwise model similarity analysis as in Figure 16, excluding the frequency and semantic diversity features (selectivity- and raw $R^2$-weighted, respectively).

\onecolumn
\begin{longtable}{p{0.22\textwidth} p{0.56\textwidth} p{0.18\textwidth}}
\caption{\textbf{Psycholinguistic features used in the analysis.}}
\label{tab:all_norms_long} \\

\hline
\textbf{Feature} & \textbf{Description} & \textbf{Category} \\
\hline
\endfirsthead

\hline
\textbf{Feature} & \textbf{Description} & \textbf{Category} \\
\hline
\endhead

\hline
\endfoot

\hline
\endlastfoot
Freq\_Blog &
Log10 version of the frequency norms based on sources from blogs. & Frequency \\
Freq\_CobS &
Log10 of word frequencies in spoken English based on COBUILD corpus. & Frequency \\
Freq\_CobW &
Log10 of word frequencies in written English based on COBUILD corpus. & Frequency \\
Freq\_HAL &
Log10 version of frequency norms based on the Hyperspace Analogue to Language (HAL) corpus. & Frequency \\
Freq\_KF &
Log10 version of frequency norms based on the Kucera and Francis corpus. & Frequency \\
Freq\_News &
Log10 version of the frequency norms based on sources from newspapers. & Frequency \\
Freq\_SUBTLEXUK &
Log10 version of the frequency norms based on SUBTLEXuk corpus. & Frequency \\
Freq\_SUBTLEXUSL &
Log10 version of frequency norms based on the SUBTLEXus corpus. & Frequency \\
Freq\_Twitter &
Log10 version of the frequency norms based on sources from Twitter. & Frequency \\
Freq\_TASA &
How experience with a word is distributed over time based on the TASA corpus. It was computed by first taking logarithms of the frequencies and then transforming them to z-values for low (first three grades) and high grades (last three grades) respectively. & Frequency \\
BOI &
The ease with which the human body can interact with a word’s referent on a scale from 1 (low interaction) to 7 (high interaction). & Motor \\
Foot\_Leg\_Lanc &
To what extent one experiences the referent by performing an action with the foot/leg, from 0 (not experienced at all) to 5 (experienced greatly). & Motor \\
Hand\_Arm\_Lanc &
To what extent one experiences the referent by performing an action with the hand/arm, from 0 (not experienced at all) to 5 (experienced greatly). & Motor \\
Head\_Lanc &
To what extent one experiences the referent by performing an action with the head, from 0 (not experienced at all) to 5 (experienced greatly). & Motor \\
Interoceptive\_Lanc &
To what extent one experiences the referent by sensations inside one’s body, from 0 (not experienced at all) to 5 (experienced greatly). & Motor \\
Mouth\_Throat\_Lanc &
To what extent one experiences the referent by performing an action with the Mouth/throat, from 0 (not experienced at all) to 5 (experienced greatly). & Motor \\
Torso\_Lanc &
To what extent one experiences the referent by performing an action with the torso, from 0 (not experienced at all) to 5 (experienced greatly). & Motor \\

Auditory\_Lanc &
To what extent one experiences the referent by hearing, from 0 (not experienced at all) to 5 (experienced greatly). & Sensory \\

Gustatory\_Lanc &
To what extent one experiences the referent by tasting, from 0 (not experienced at all) to 5 (experienced greatly). & Sensory \\

Haptic\_Lanc &
To what extent one experiences the referent by feeling through touch, from 0 (not experienced at all) to 5 (experienced greatly). & Sensory \\

Olfactory\_Lanc &
To what extent one experiences the referent by smelling, from 0 (not experienced at all) to 5 (experienced greatly). & Sensory \\

Sensory\_Experience &
The extent to which a word evokes a sensory and/or perceptual experience in the mind of the reader on a 1 to 7 scale, with higher numbers indicating a greater sensory experience. & Sensory \\

Visual\_Lanc &
To what extent one experiences the referent by seeing, from 0 (not experienced at all) to 5 (experienced greatly). & Sensory \\

CD\_Blog &
Log10 version of the contextual diversity of a word, which refers to the number of passages (documents) in the sources from Blog containing a particular word. & Semantic diversity \\

CD\_News &
Log10 version of the contextual diversity of a word, which refers to the number of passages (documents) in the sources from newspapers containing a particular word. & Semantic diversity \\

CD\_SUBTLEXUK &
Log10 version of the contextual diversity of a word, which refers to the number of passages (documents) in the SUBTLEXuk corpus containing a particular word. & Semantic diversity \\

CD\_SUBTLEXUS &
Log10 version of the contextual diversity of a word, which refers to the number of passages (documents) in the SUBTLEXus corpus containing a particular word. & Semantic diversity \\

CD\_Twitter &
Log10 version of the contextual diversity of a word, which refers to the number of passages (documents) in the sources from Twitter containing a particular word. & Semantic diversity \\

Sem\_Diversity &
The degree to which different contexts associated with a word vary in their meanings. & Semantic diversity \\

LexicalD\_ACC\_V\_BLP &
The proportion of accurate responses of visual lexical decision for a particular word from the British Lexicon Project. & Visual lexical decision \\

LexicalD\_ACC\_V\_ECPP &
The proportion of accurate responses of visual lexical decision for a particular word from the English Crowdsourcing Project. & Visual lexical decision \\

LexicalD\_ACC\_V\_ELP &
The proportion of accurate responses of visual lexical decision for a particular word from the English Lexicon Project. & Visual lexical decision \\

LexicalD\_RT\_V\_BLP &
The mean visual lexical decision latency (in msec) for a particular word across participants from the British Lexicon Project. & Visual lexical decision \\

LexicalD\_RT\_V\_ECP &
The mean visual lexical decision latency (in msec) for a particular word in the word knowledge task across participants from the English Crowdsourcing Project. This task is similar, but not identical, to the traditional lexical decision task. Participants were asked to indicate whether each item “is a word you know or not.” Their results showed that RTs in this task correlate well with those from lexical decision in ELP and BLP, and hence we have labelled it as such. & Visual lexical decision \\

LexicalD\_RT\_V\_ELP &
The mean visual lexical decision latency (in msec) for a particular word across participants from the English Lexicon Project. & Visual lexical decision \\

LexicalD\_ACC\_A\_AELP&
The proportion of accurate responses of auditory lexical decision for a particular word from the Auditory English Lexicon Project. & Auditory lexical decision \\

LexicalD\_ACC\_A\_MALD &
The proportion of accurate responses of auditory lexical decision for a particular word from the Massive Auditory Lexical Decision database. & Auditory lexical decision \\

LexicalD\_RT\_A\_AELP&
The mean auditory lexical decision latency (in msec) for a particular word from the Auditory English Lexicon Project. & Auditory lexical decision \\

LexicalD\_RT\_A\_MALD &
The mean auditory lexical decision latency (in msec) for a particular word from the Massive Auditory Lexical Decision database.& Auditory lexical decision \\

AoA\_Kuper &
The age at which people acquired the word, in which  participants were asked to enter the age (in years) at which they thought they had learned the word.& Familiarity \\

Fam\_Brys &
Percentage of participants who know the word well enough to give answer for the concreteness rating. & Familiarity \\

Prevalence\_Brys &
The proportion of people who know the word, in which participants were asked to indicate whether or not they knew the stimulus in a list of words and nonwords, in an online crowdsourcing study. Percentages were translated to z values on the the basis of cumulative normal distribution. & Familiarity \\

perc\_known\_winter &
Percentage of participants that did not know the meaning or pronunciation of the word. & Familiarity \\

Valence\_NRC &
Word-emotion association built by manual annotation using Best-Worst Scaling method, with scores ranging from 0 (negative) to 1 (positive). & Valence \\

Valence\_Warr &
The pleasantness of a stimulus on a 1 (happy) to 9 (unhappy) scale. & Valence \\

Arousal\_NRC &
Word-emotion association built by manual annotation using Best-Worst Scaling method, with scores ranging from 0 (low arousal) to 1 (high arousal). & Arousal \\

Arousal\_Warr &
The intensity of emotion provoked by a stimulus on a 1 (aroused) to 9 (calm) scale. & Arousal \\

Dominance\_NRC &
Word-emotion association built by manual annotation using Best-Worst Scaling method, with scores ranging from 0 (low dominance) to 1 (high dominance). & Dominance \\

Dominance\_Warr &
The degree of control exerted by a stimulus on a 1 (controlled) to 9 (in control) scale. & Dominance \\

Naming\_ACC\_ELP &
The proportion of accurate responses of word naming for a particular word from the English Lexicon Project. & Naming \\

Naming\_RT\_ELP &
The mean naming latency (in msec) for a particular word across participants from the English Lexicon Project. & Naming \\

SemanticD\_ACC\_Calgary &
The proportion of accurate responses of concrete/abstract semantic decision (i.e., does the word refer to something concrete or abstract?) for a particular word from the Calgary database. & Semantic decision \\

SemanticD\_RT\_Calgary &
The mean latency (in msec) of concrete/abstract semantic decision (i.e., does the word refer to something concrete or abstract?) for a particular word from the Calgary database. & Semantic decision \\

Sem\_N\_D &
The average radius of co-occurrence, which is the average distance between the words in the semantic neighborhood and the target word. & Semantic neighborhood \\

AoA\_LWV &
The age at which people acquired the word, in which a three-choice test was administered to participants in grades 4 to 16 (college) (Living Word Vocabulary database). & Age of acquisition \\

Conc\_Brys &
The degree to which the concept can be experienced directly through the senses from a 1 (abstract) to 5 (concrete) scale. & Concreteness \\

Socialness &
The extent to which a word's meaning has social relevance on a seven-point Likert scale from 1 to 7. & Social/moral \\

iconicity\_winter\_2023 &
Iconicty ratings on a scale from 0 (not iconic at all) to 7 (very iconic). & Iconicity/transparency \\

\end{longtable}
\twocolumn

\begin{figure*}[h!]
    \centering
    \includegraphics[width=\textwidth]{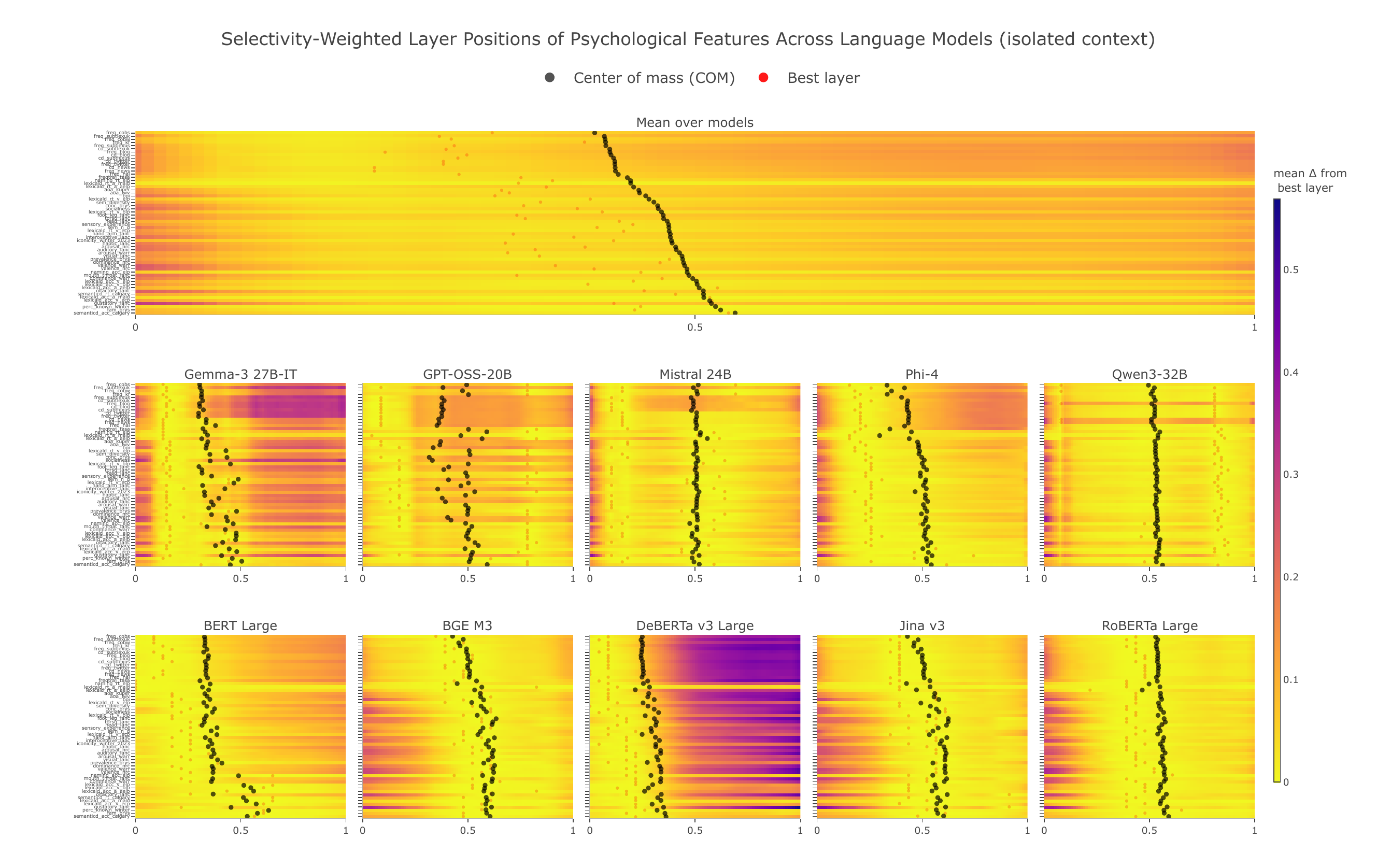}
  \caption{Selectivity-weighted layer positions of psycholinguistic features in the isolated context.
Each panel shows a language model (decoders top row, encoders bottom row). The heatmap depicts the $\Delta$-from-best-layer across layers for each feature (x-axis: normalized layer index from first to last; y-axis: psycholinguistic features) based on Selectivity score. Black dots indicate the Selectivity-weighted center of mass (COM) of each feature’s layer profile, while red dots mark the single best-performing layer (argmax). Lower (yellow) heatmap values indicate layers closer to the optimal representation of a feature.}
  \label{fig:norm-isolated-selectivity}
\end{figure*}

\begin{figure*}[h!]
    \centering
  \includegraphics[width=\textwidth]{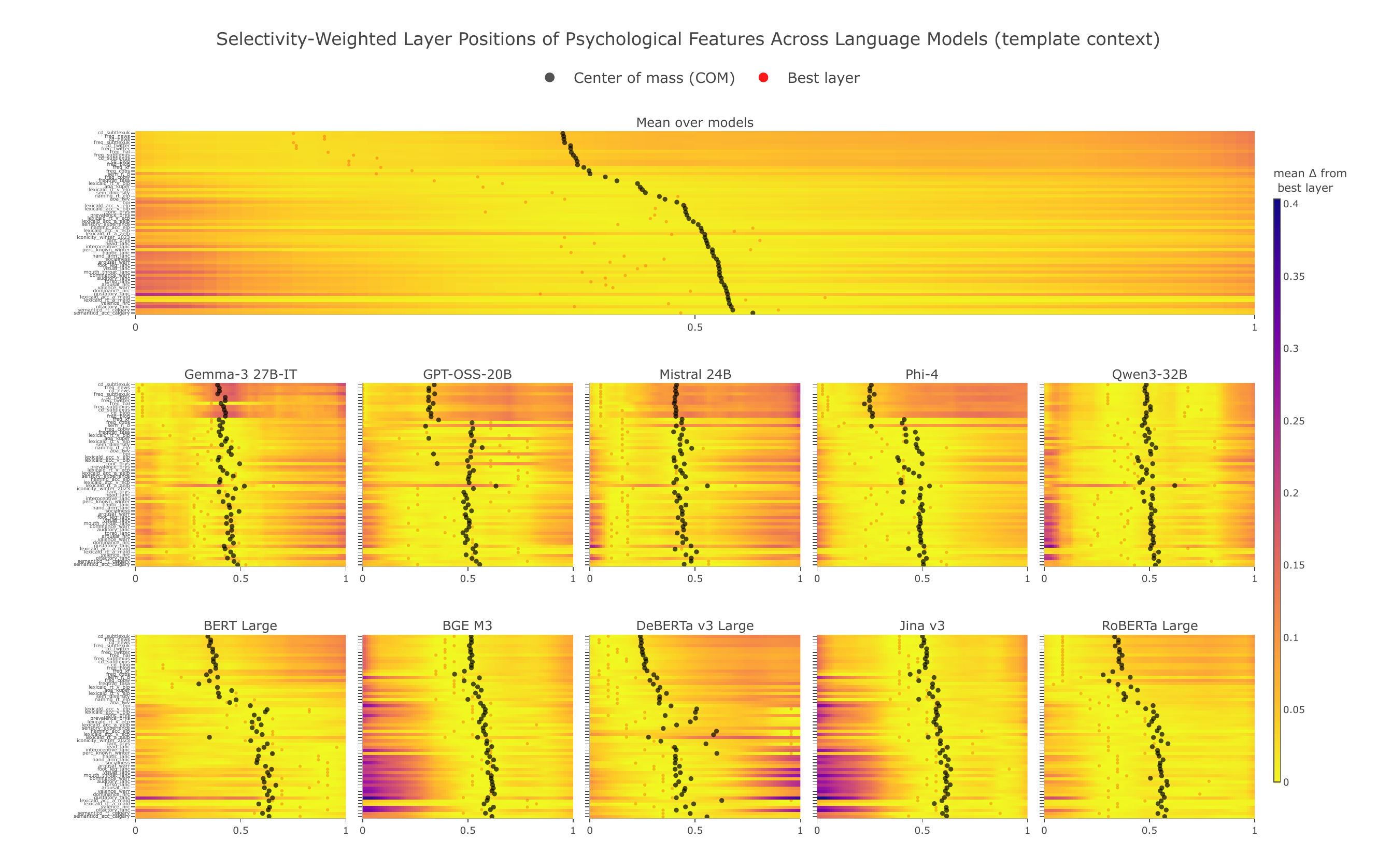}
  \caption{Selectivity-weighted layer positions of psycholinguistic features in the template context.
Each panel shows a language model (decoders top row, encoders bottom row). The heatmap depicts the $\Delta$-from-best-layer across layers for each feature (x-axis: normalized layer index from first to last; y-axis: psycholinguistic features) based on Selectivity score. Black dots indicate the Selectivity-weighted center of mass (COM) of each feature’s layer profile, while red dots mark the single best-performing layer (argmax). Lower (yellow) heatmap values indicate layers closer to the optimal representation of a feature.}
  \label{fig:norm-template-selectivity}
\end{figure*}

\begin{figure*}[h!]
    \centering
  \includegraphics[width=\textwidth]{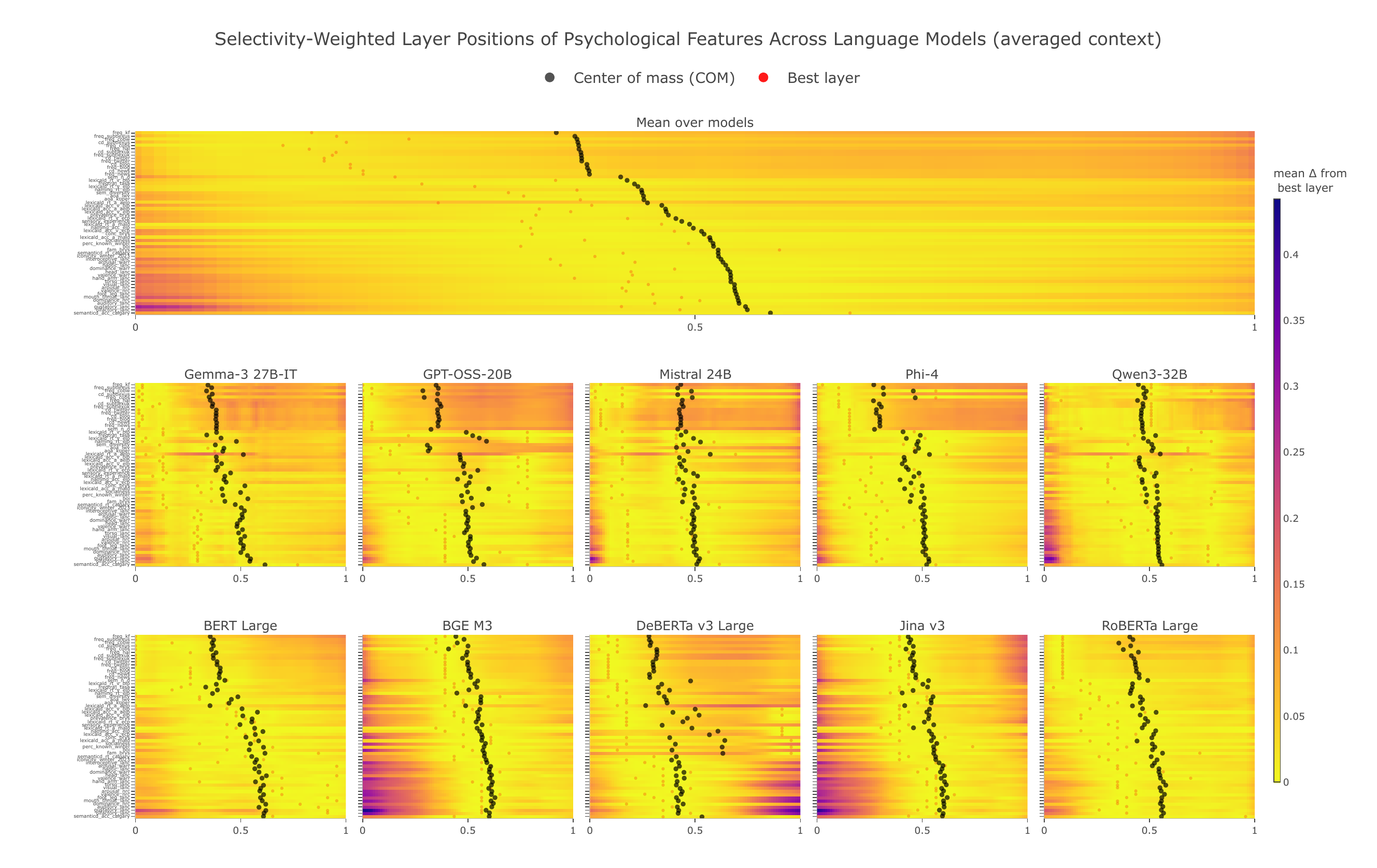}
  \caption{Selectivity-weighted layer positions of psycholinguistic features in the averaged context.
Each panel shows a language model (decoders top row, encoders bottom row). The heatmap depicts the $\Delta$-from-best-layer across layers for each feature (x-axis: normalized layer index from first to last; y-axis: psycholinguistic features) based on Selectivity score. Black dots indicate the Selectivity-weighted center of mass (COM) of each feature’s layer profile, while red dots mark the single best-performing layer (argmax). Lower (yellow) heatmap values indicate layers closer to the optimal representation of a feature.}
  \label{fig:norm-averaged-raw}
\end{figure*}

\begin{figure*}[h!]
    \centering
  \includegraphics[width=\textwidth]{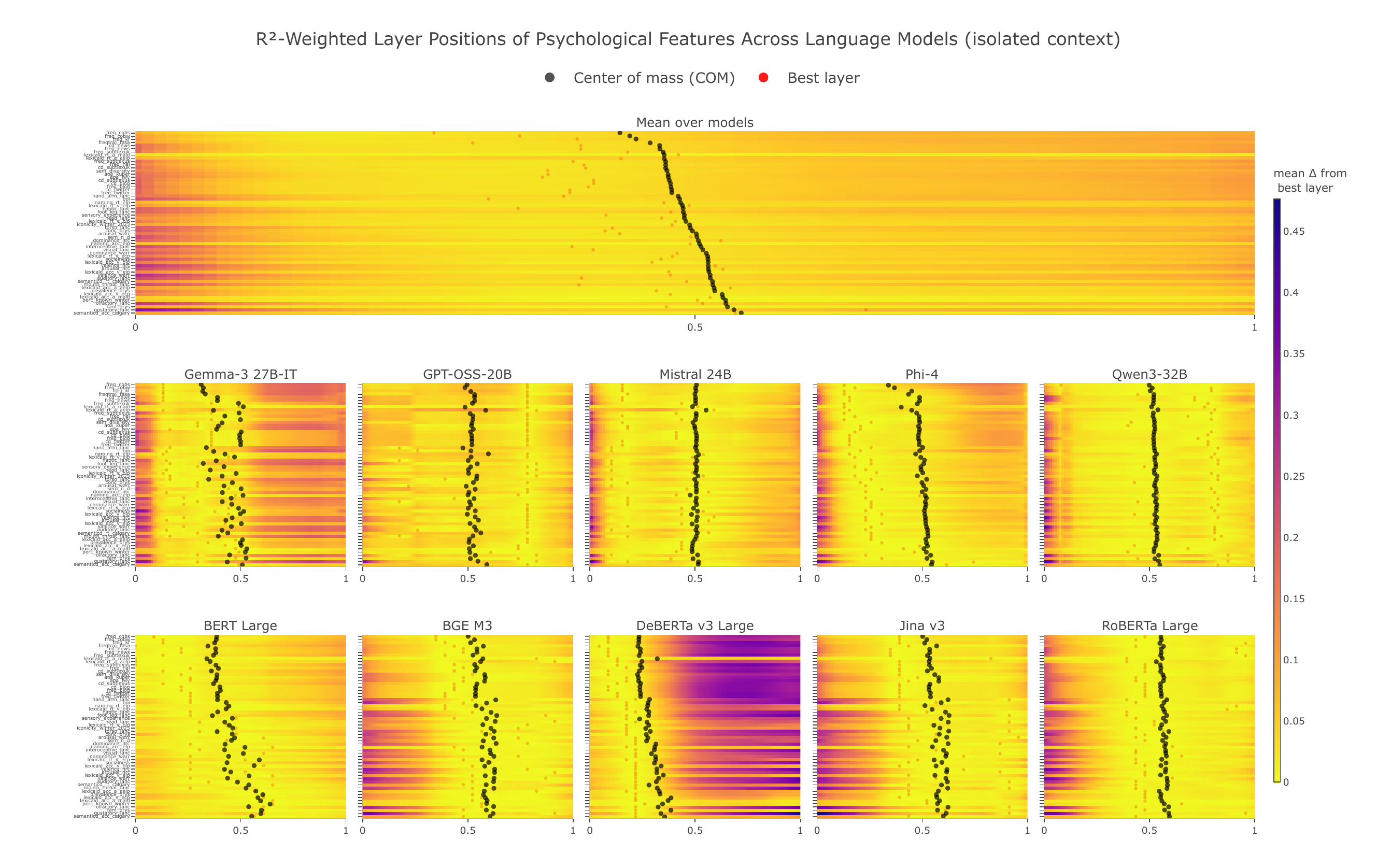}
  \caption{Raw $R^2$-weighted layer positions of psycholinguistic features in the isolated context.
Each panel shows a language model (decoders top row, encoders bottom row). The heatmap depicts the $\Delta$-from-best-layer across layers for each feature (x-axis: normalized layer index from first to last; y-axis: psycholinguistic features) based on Raw $R^2$ score. Black dots indicate the Raw $R^2$-weighted center of mass (COM) of each feature’s layer profile, while red dots mark the single best-performing layer (argmax). Lower (yellow) heatmap values indicate layers closer to the optimal representation of a feature.}
  \label{fig:norm-isolated-raw}
\end{figure*}

\begin{figure*}[h!]
    \centering
  \includegraphics[width=\textwidth]{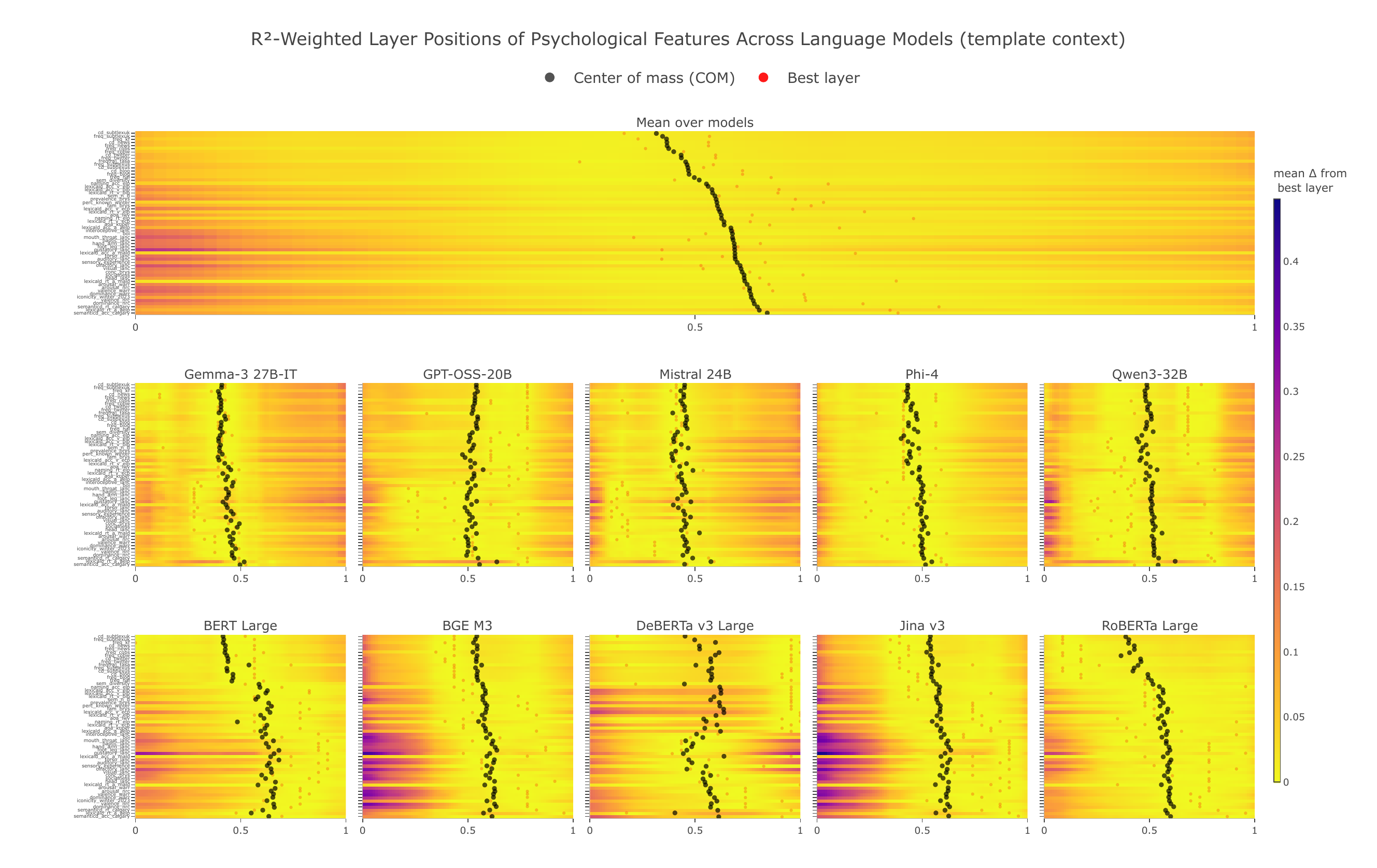}
  \caption{Raw $R^2$-weighted layer positions of psycholinguistic features in the template context.
Each panel shows a language model (decoders top row, encoders bottom row). The heatmap depicts the $\Delta$-from-best-layer across layers for each feature (x-axis: normalized layer index from first to last; y-axis: psycholinguistic features) based on Raw $R^2$ score. Black dots indicate the Raw $R^2$-weighted center of mass (COM) of each feature’s layer profile, while red dots mark the single best-performing layer (argmax). Lower (yellow) heatmap values indicate layers closer to the optimal representation of a feature.}
  \label{fig:norm-template-raw}
\end{figure*}

\begin{figure*}[h!]
    \centering
  \includegraphics[width=\textwidth]{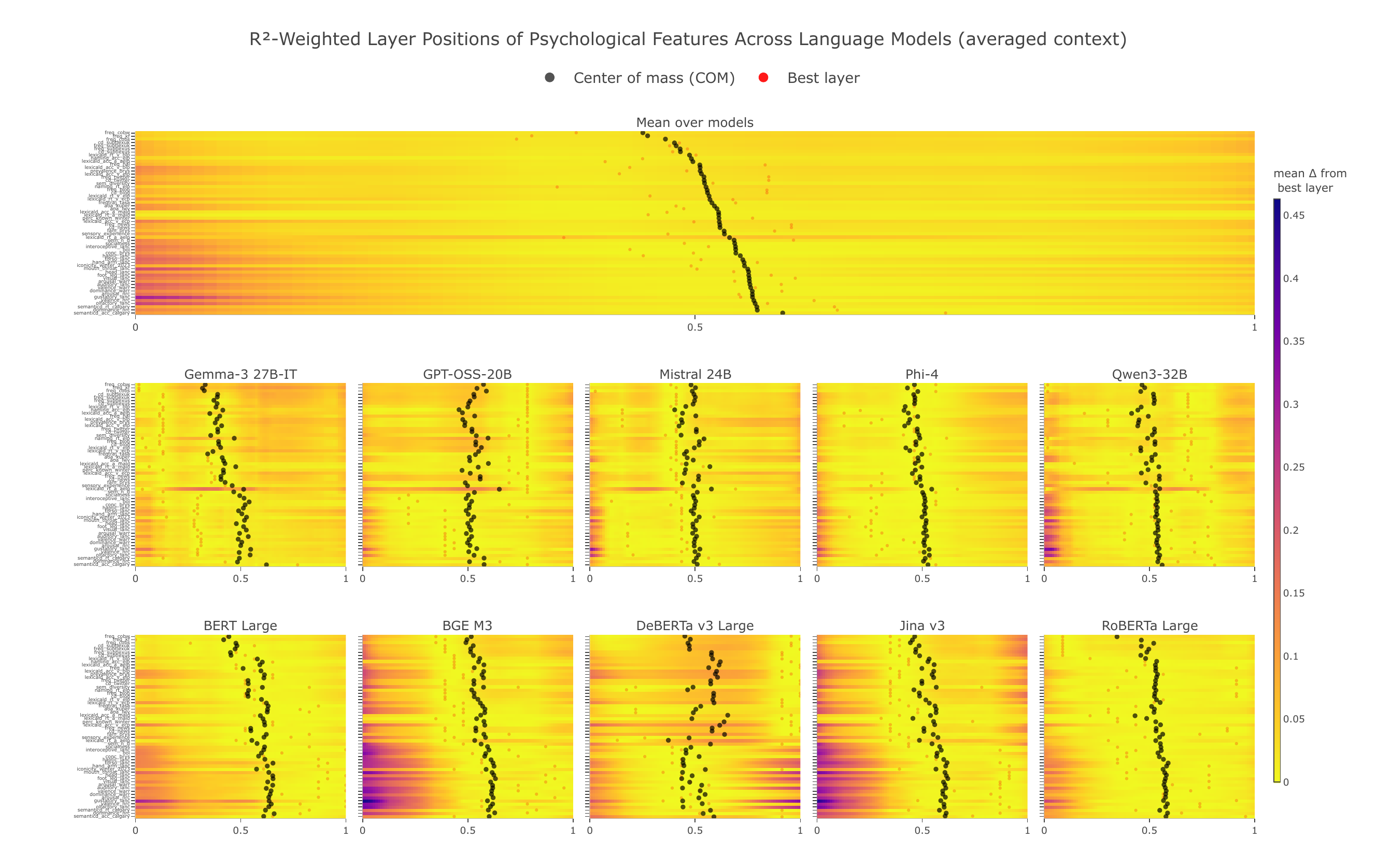}
  \caption{Raw $R^2$-weighted layer positions of psycholinguistic features in the averaged context.
Each panel shows a language model (decoders top row, encoders bottom row). The heatmap depicts the $\Delta$-from-best-layer across layers for each feature (x-axis: normalized layer index from first to last; y-axis: psycholinguistic features) based on Raw $R^2$ score. Black dots indicate the Raw $R^2$-weighted center of mass (COM) of each feature’s layer profile, while red dots mark the single best-performing layer (argmax). Lower (yellow) heatmap values indicate layers closer to the optimal representation of a feature.}
  \label{fig:norm-averaged-raw}
\end{figure*}

\begin{figure*}[h!]
    \centering
  \includegraphics[width=\textwidth]{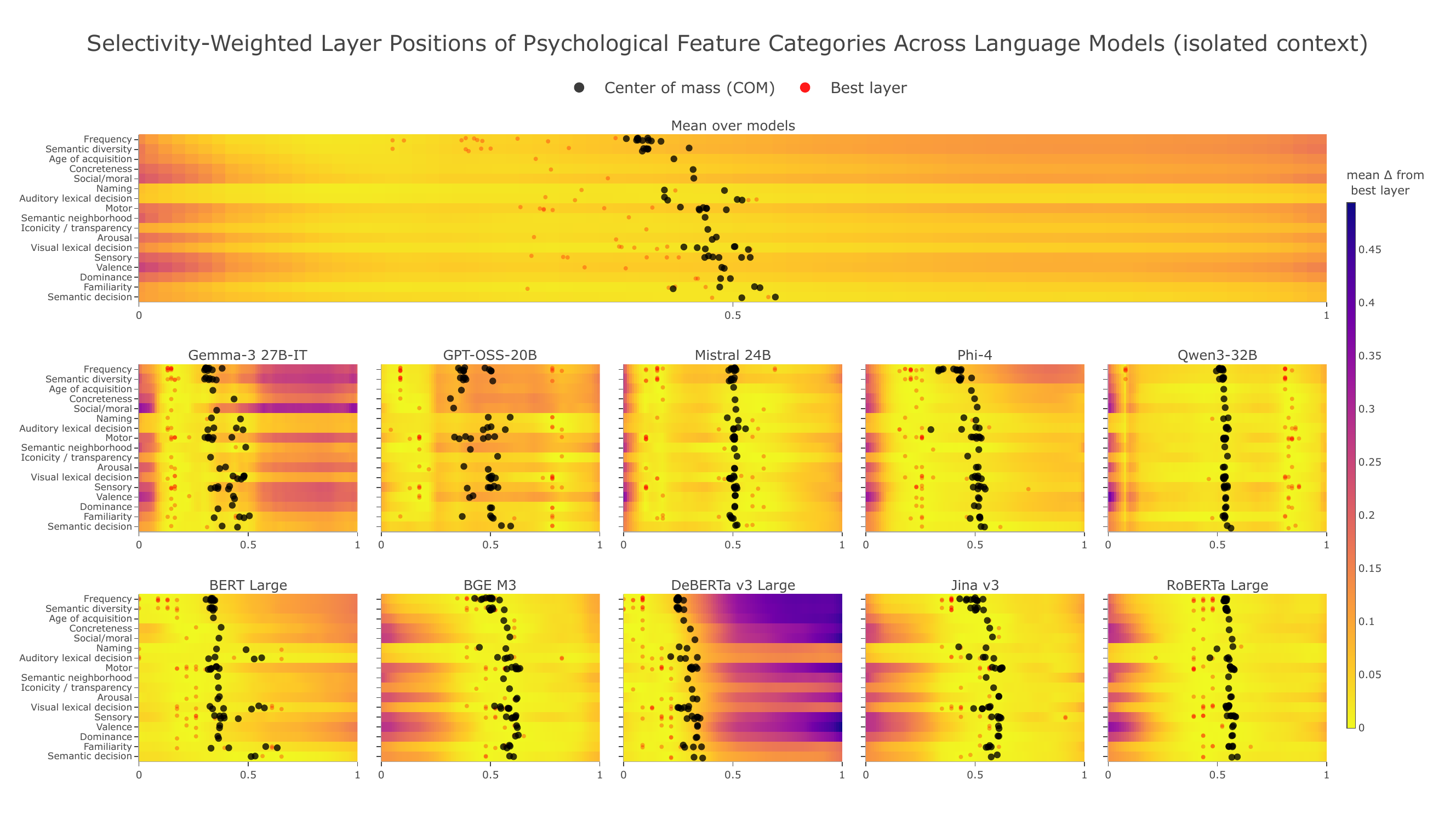}
  \caption{Selectivity-weighted layer positions of psycholinguistic feature categories in the isolated context.
Each panel shows a language model (decoders top row, encoders bottom row). The heatmap depicts the mean $\Delta$-from-best-layer across features comprising each category (x-axis: normalized layer index from first to last; y-axis: psycholinguistic feature categories) based on Selectivity score. Black dots indicate the Selectivity-weighted center of mass (COM) of each category’s layer profile, while red dots mark the single best-performing layer (argmax). Lower (yellow) heatmap values indicate layers closer to the optimal representation of a category.}
  \label{fig:isolated-selectivity}
\end{figure*}

\begin{figure*}[h!]
    \centering
  \includegraphics[width=\textwidth]{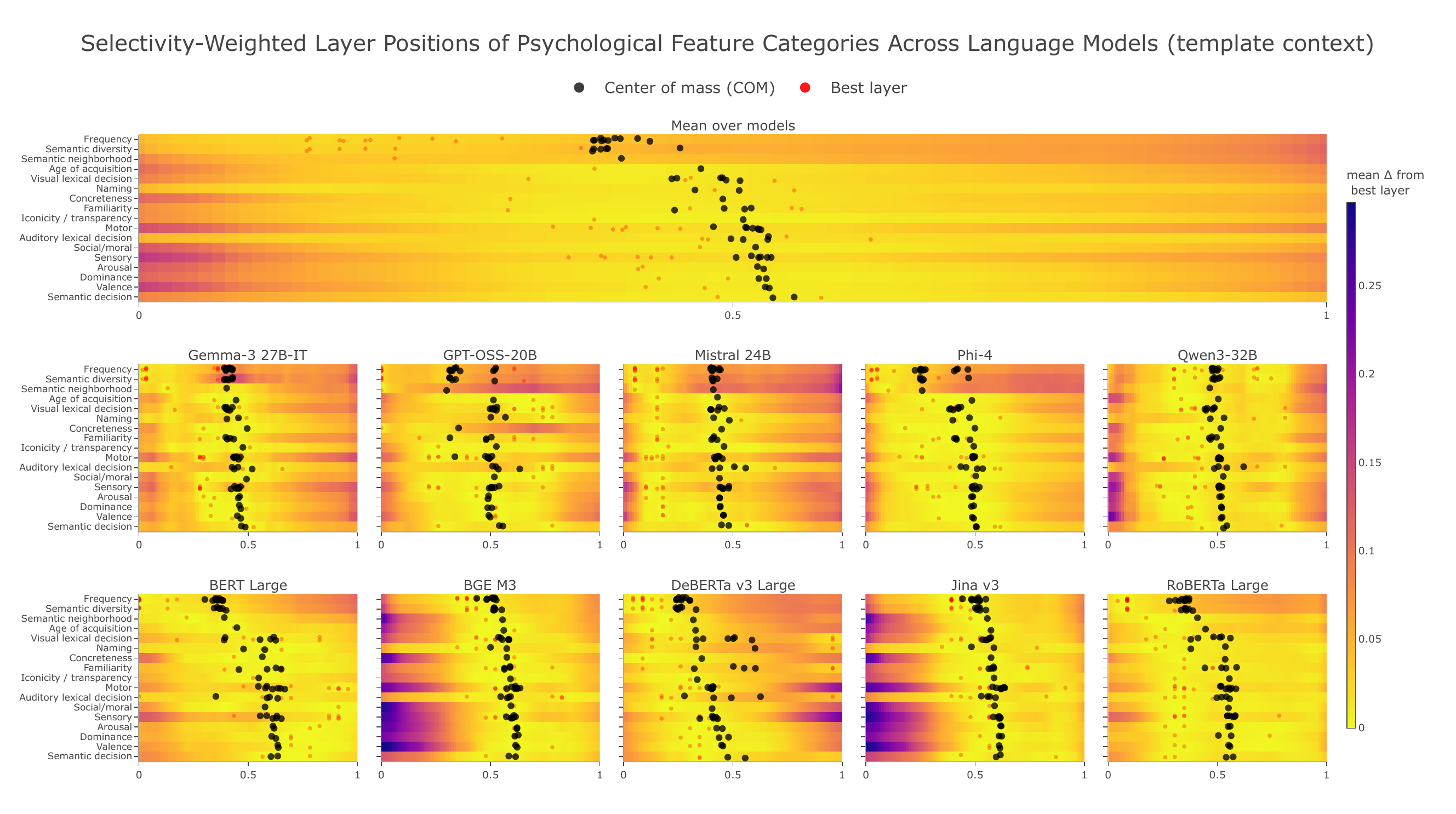}
  \caption{Selectivity-weighted layer positions of psycholinguistic feature categories in the template context.
Each panel shows a language model (decoders top row, encoders bottom row). The heatmap depicts the mean $\Delta$-from-best-layer across features comprising each category (x-axis: normalized layer index from first to last; y-axis: psycholinguistic feature categories) based on Selectivity score. Black dots indicate the Selectivity-weighted center of mass (COM) of each category’s layer profile, while red dots mark the single best-performing layer (argmax). Lower (yellow) heatmap values indicate layers closer to the optimal representation of a category.}
  \label{fig:template-selectivity}
\end{figure*}

\begin{figure*}[h!]
    \centering
  \includegraphics[width=\textwidth]{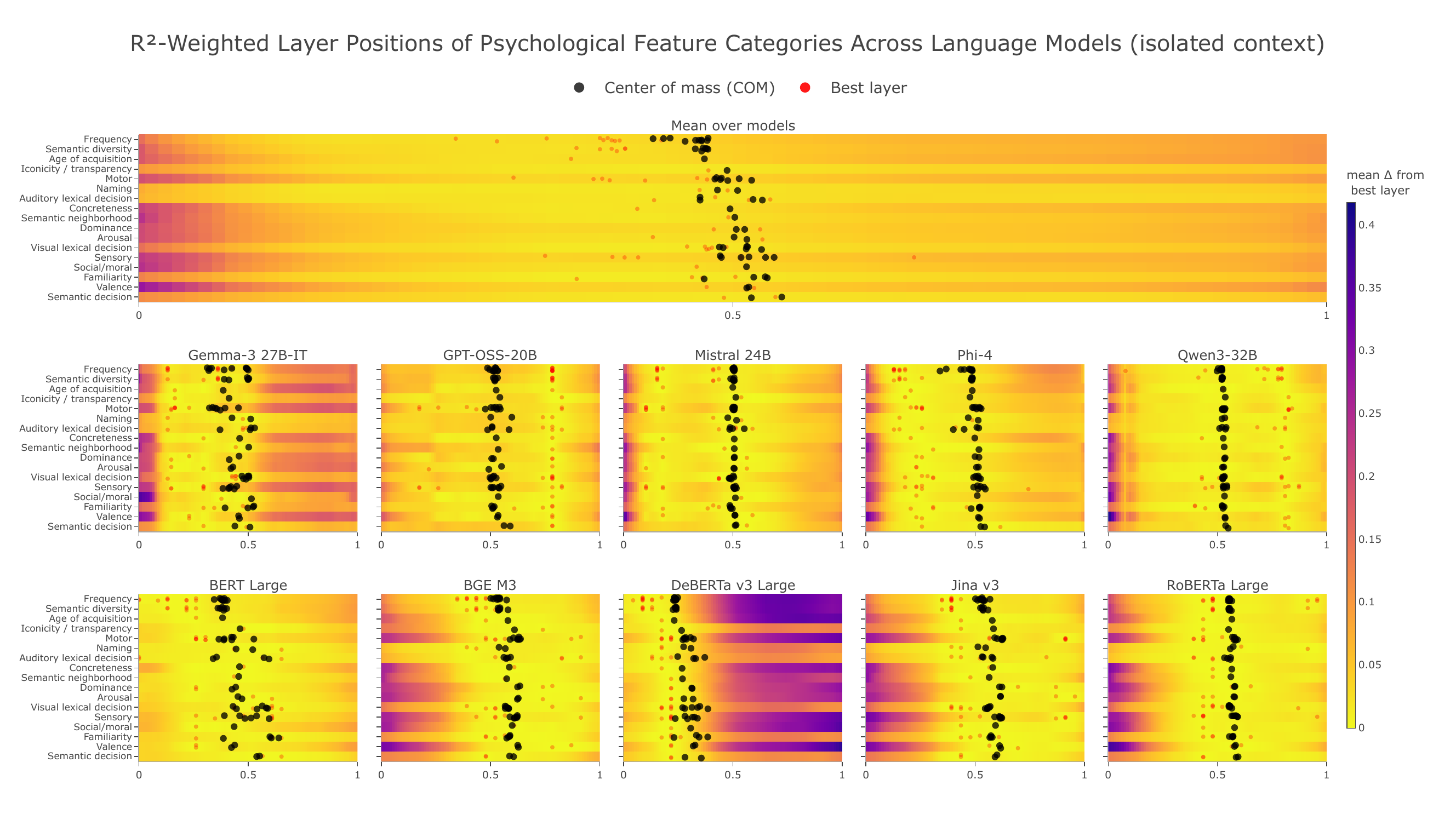}
  \caption{Raw $R^2$-weighted layer positions of psycholinguistic feature categories in the isolated context.
Each panel shows a language model (decoders top row, encoders bottom row). The heatmap depicts the mean $\Delta$-from-best-layer across features comprising each category (x-axis: normalized layer index from first to last; y-axis: psycholinguistic feature categories) based on Raw $R^2$ score. Black dots indicate the Raw $R^2$-weighted center of mass (COM) of each category’s layer profile, while red dots mark the single best-performing layer (argmax). Lower (yellow) heatmap values indicate layers closer to the optimal representation of a category.}
  \label{fig:isolated-raw}
\end{figure*}

\begin{figure*}[h!]
    \centering
  \includegraphics[width=\textwidth]{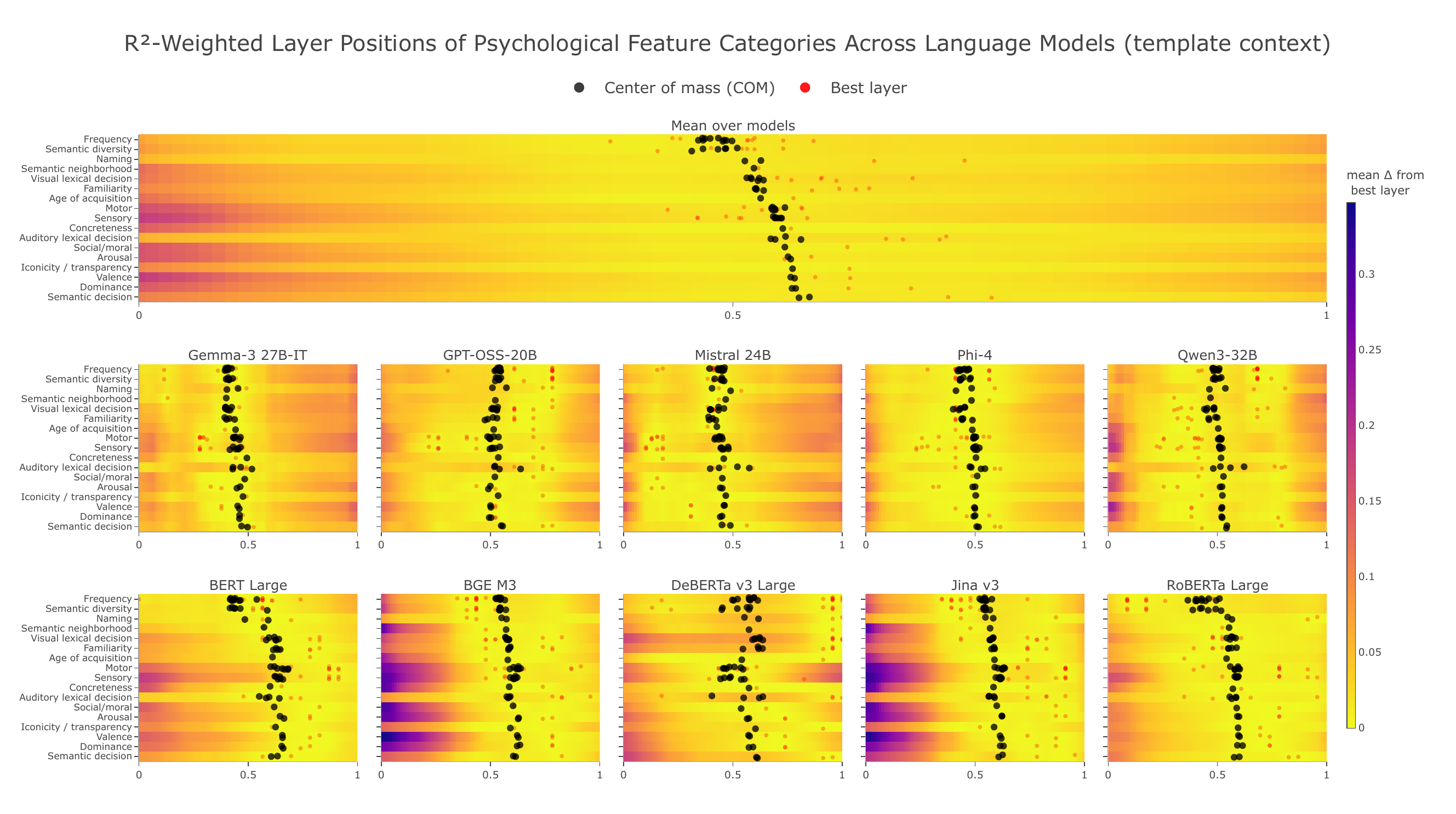}
  \caption{Raw $R^2$-weighted layer positions of psycholinguistic feature categories in the template context.
Each panel shows a language model (decoders top row, encoders bottom row). The heatmap depicts the mean $\Delta$-from-best-layer across features comprising each category (x-axis: normalized layer index from first to last; y-axis: psycholinguistic feature categories) based on Raw $R^2$ score. Black dots indicate the Raw $R^2$-weighted center of mass (COM) of each category’s layer profile, while red dots mark the single best-performing layer (argmax). Lower (yellow) heatmap values indicate layers closer to the optimal representation of a category.}
  \label{fig:template-raw}
\end{figure*}

\begin{figure*}[h!]
    \centering
  \includegraphics[width=\textwidth]{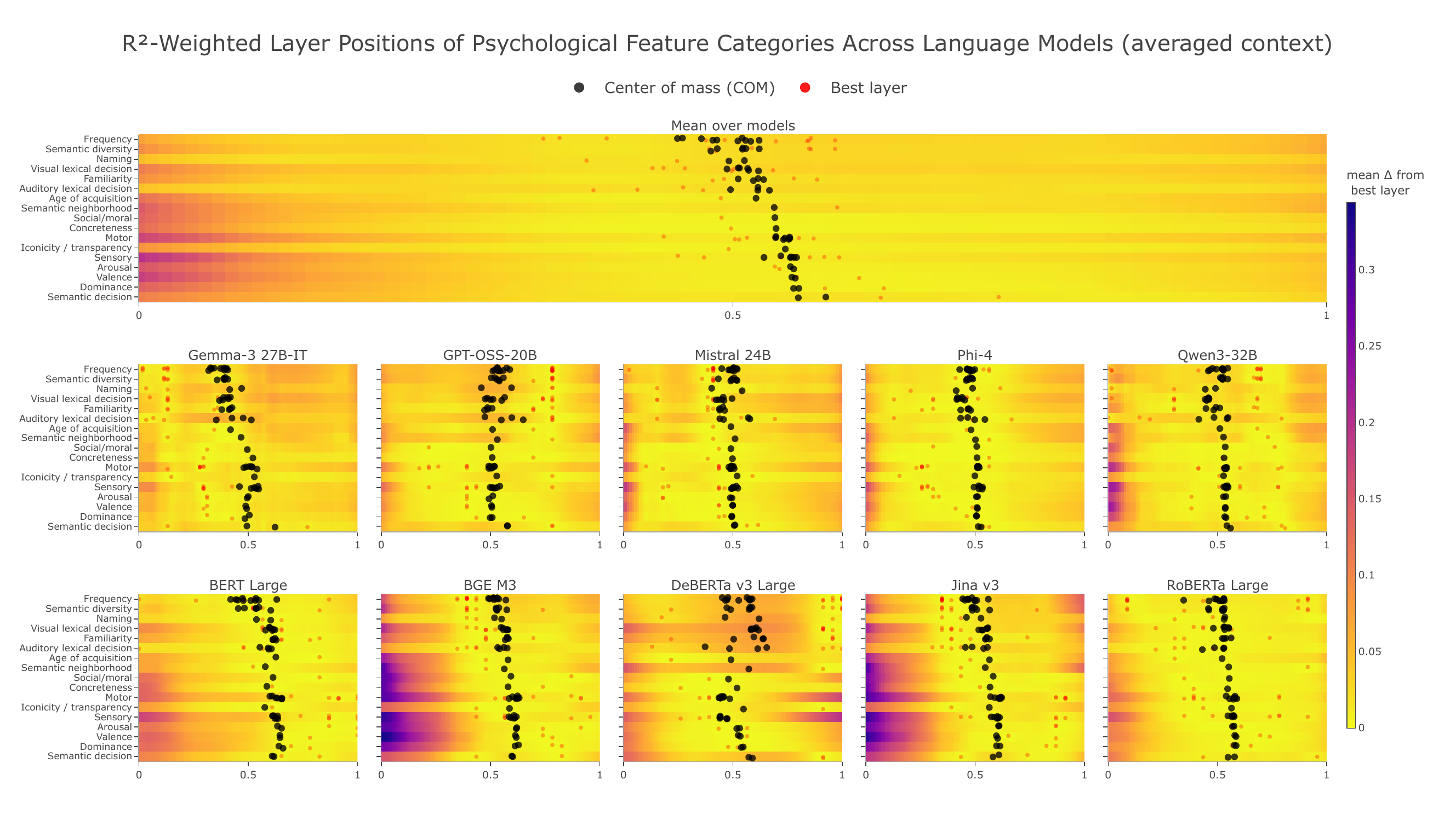}
  \caption{Raw $R^2$-weighted layer positions of psycholinguistic feature categories in the averaged context.
Each panel shows a language model (decoders top row, encoders bottom row). The heatmap depicts the mean $\Delta$-from-best-layer across features comprising each category (x-axis: normalized layer index from first to last; y-axis: psycholinguistic feature categories) based on Raw $R^2$ score. Black dots indicate the Raw $R^2$-weighted center of mass (COM) of each category’s layer profile, while red dots mark the single best-performing layer (argmax). Lower (yellow) heatmap values indicate layers closer to the optimal representation of a category.}
  \label{fig:averaged_raw}
\end{figure*}

\begin{figure*}[h!]
    \centering
  \includegraphics[width=\textwidth]{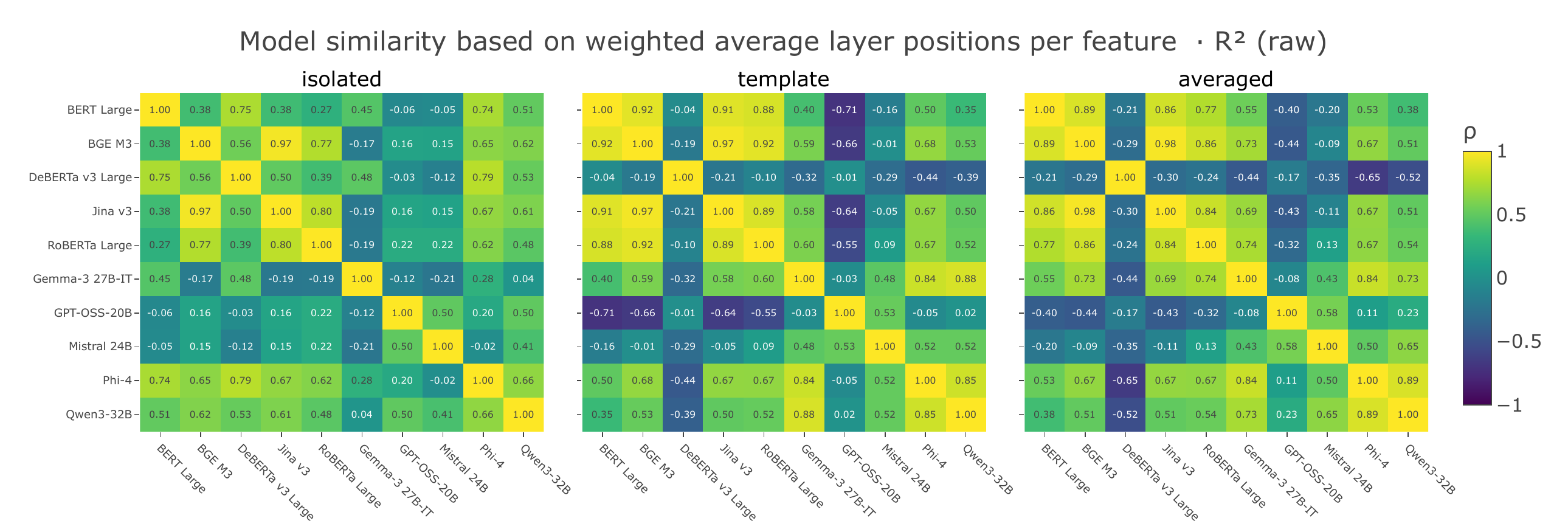}
  \caption{Each panel shows the pairwise Spearman correlation (p) between models, computed from vectors of feature-specific center-of-mass (COM) layer positions within a given context (isolated, template, averaged). For each model and feature, the COM is calculated as the score-weighted mean of normalized layer indices (using raw $R^2$ scores), summarizing where in the network a feature is most strongly represented. Correlations are computed across features, yielding a similarity matrix that reflects how similarly different models localize psycholinguistic information across layers.}
  \label{fig:corr_raw}
\end{figure*}

\begin{figure*}[h!]
    \centering
  \includegraphics[width=\textwidth]{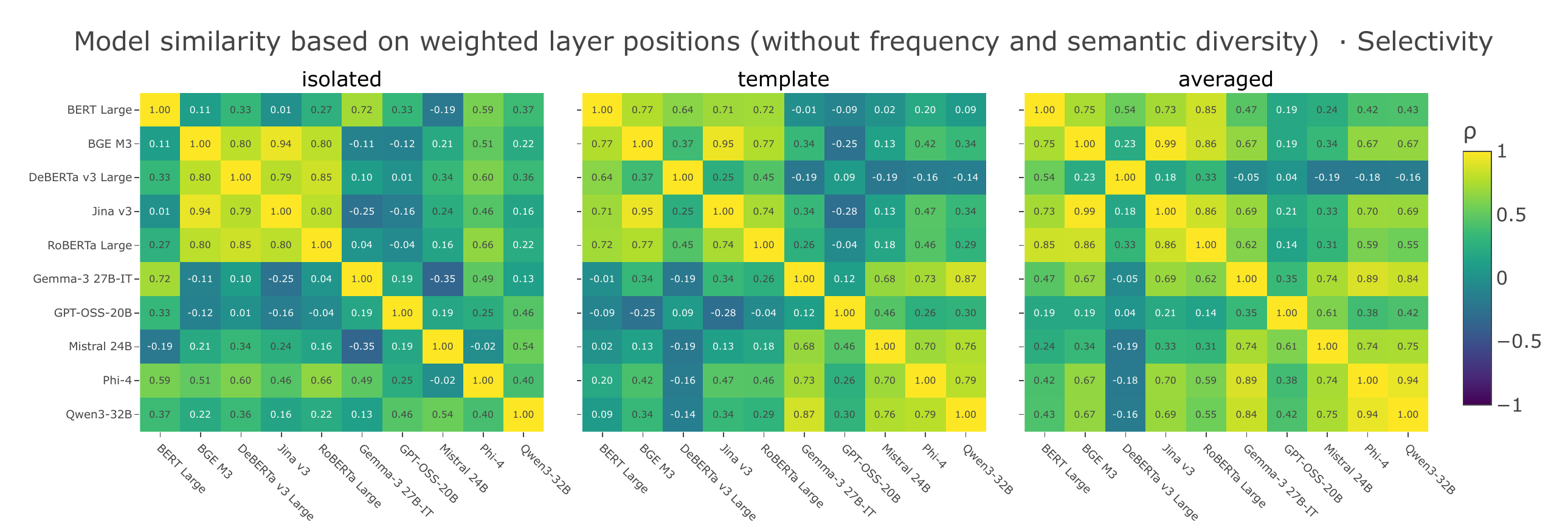}
  \caption{Each panel shows the pairwise Spearman correlation (p) between models, computed from vectors of feature-specific center-of-mass (COM) layer positions within a given context (isolated, template, averaged) excluding frequency and semantic diversity features. For each model and feature, the COM is calculated as the score-weighted mean of normalized layer indices (using selectivity scores), summarizing where in the network a feature is most strongly represented. Correlations are computed across features, yielding a similarity matrix that reflects how similarly different models localize psycholinguistic information across layers.}
  \label{fig:filtered_corr_selectivity}
\end{figure*}

\begin{figure*}[h!]
    \centering
  \includegraphics[width=\textwidth]{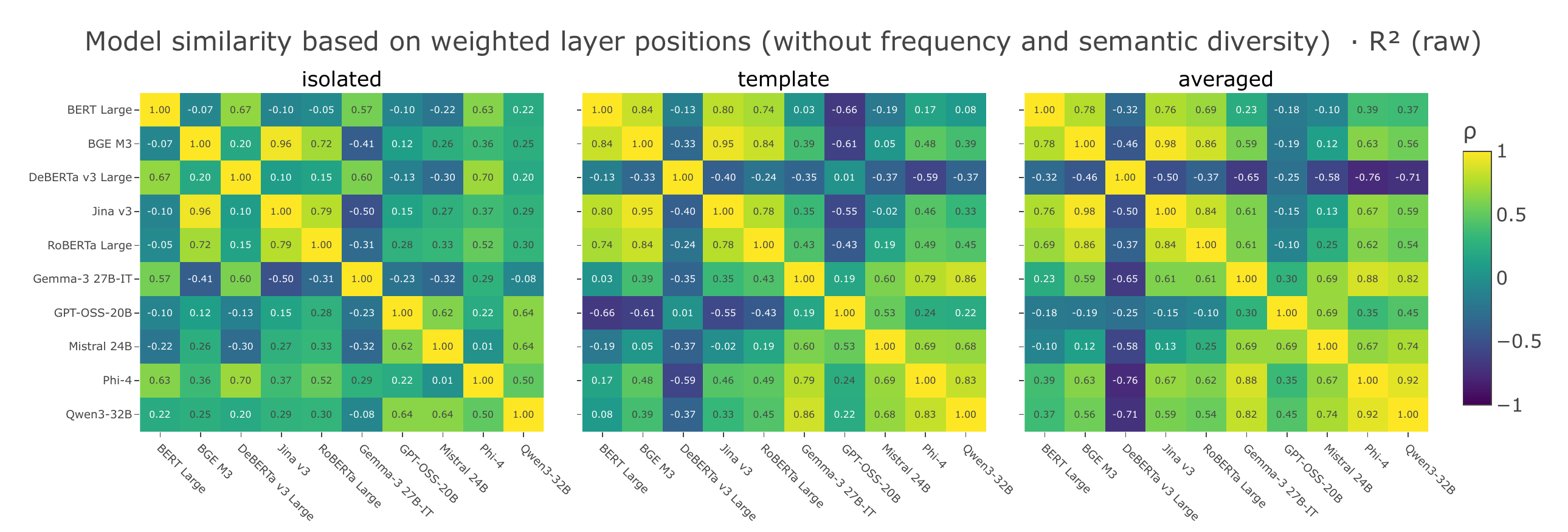}
  \caption{Each panel shows the pairwise Spearman correlation (p) between models, computed from vectors of feature-specific center-of-mass (COM) layer positions within a given context (isolated, template, averaged) excluding frequency and semantic diversity features. For each model and feature, the COM is calculated as the score-weighted mean of normalized layer indices (using raw $R^2$ scores), summarizing where in the network a feature is most strongly represented. Correlations are computed across features, yielding a similarity matrix that reflects how similarly different models localize psycholinguistic information across layers.}
  \label{fig:filtered_corr_raw}
\end{figure*}

\end{document}